\documentclass[twocolumn,preprint]{elsarticle}

\usepackage{lineno,hyperref}
\usepackage{url}
\usepackage{hyperref}
\usepackage{times}
\usepackage{epsfig}
\usepackage{graphicx}
\usepackage{amsmath}
\usepackage{multirow}
\usepackage{amssymb}
\usepackage{subfig}
\usepackage{graphicx}
\usepackage{float}
\usepackage{booktabs}
\usepackage{textcomp}
\modulolinenumbers[255]

\journal{Neurocomputing}

\begin{document}

\twocolumn[{
\begin{frontmatter}

\title{SuperCon: Supervised Contrastive Learning for Imbalanced Skin Lesion Classification}

\author{Keyu Chen, Di Zhuang, J. Morris Chang\\
{\tt\small \{keyu, dizhuang, chang5\}@usf.edu}\\}
\address{Department of Electrical Engineering, University of South Florida, Tampa, FL 33620}

\begin{abstract} 

Convolutional neural networks (CNNs) have achieved great success in skin lesion classification. A balanced dataset is required to train a good model. However, due to the appearance of different skin lesions in practice, severe or even deadliest skin lesion types (e.g., melanoma) naturally have quite small amount represented in a dataset. In that, classification performance degradation occurs widely, it is significantly important to have CNNs that work well on class imbalanced skin lesion image dataset. In this paper, we propose SuperCon, a two-stage training strategy to overcome the class imbalance problem on skin lesion classification. It contains two stages: (i) representation training that tries to learn a feature representation that closely aligned among intra-classes and distantly apart from inter-classes, and (ii) classifier fine-tuning that aims to learn a classifier that correctly predict the label based on the learnt representations. In the experimental evaluation, extensive comparisons have been made among our approach and other existing approaches on skin lesion benchmark datasets. The results show that our two-stage training strategy effectively addresses the class imbalance classification problem, and significantly improves existing works in terms of F1-score and AUC score, resulting in state-of-the-art performance.

\end{abstract}

\begin{keyword}
\texttt{Skin image analysis, Deep learning, Class imbalance, Representation learning}
\end{keyword}
\end{frontmatter}
}]

\linenumbers

\section{Introduction}\label{sec:introduction}
Deep learning has been extensively applied in the field of skin image analysis. For example, Zhang et al. \cite{zhang2019attention} has shown that convolutional neural networks (CNNs) have achieved the state-of-the-art performance in skin image classification. The performance of a deep neural network models on skin image analysis highly depends on the quality and quantity of the dataset. However, since in most of the real-world skin image datasets \cite{gutman2016skin, codella2018skin, codella2019skin} the data of certain classes (e.g., the benign lesions) is abundant which makes them an over-represented majority, while the data of some other classes  (e.g., the cancerous lesions) is deficient which makes them an underrepresented minority. 
It is more crucial to precisely identify the samples from an underrepresented (i.e., in terms of the amount of data) but more important minority class (e.g., cancerous skin lesions). 
For instance, a deadly cancerous skin lesion (e.g., melanoma) that rarely appears during the examinations should be barely misclassified as benign or other less severe lesions (e.g., dermatofibroma). Hence, it is of great importance to enable CNNs to work well on imbalanced skin image datasets, especially for those minority but deadliest skin diseases, e.g., Melanoma.

To date, there are a few methods attempting to enhance the performance of CNNs on class imbalanced datasets \cite{hensman2015impact,lee2016plankton,pouyanfar2018dynamic, wang2016training, khan2017cost, lin2017focal, wang2018predicting, zhang2016training, zhang2018image, ren2018multi, garcia2018dynamic, cruz2015meta, zhuang2020cs}. In terms of the attention of a deep learning pipeline, the current methods can be roughly divided into three categories: data-level methods, loss-level methods, and model-level methods. 
In data-level methods, such as random over-sampling (ROS) \cite{hensman2015impact}, random under-sampling (RUS) \cite{lee2016plankton} and dynamic sampling \cite{pouyanfar2018dynamic}, techniques are usually proposed to directly adjust the training data distributions by tuning the sampling rate of the data of different classes. For instance, RUS \cite{lee2016plankton} pre-trains a model by randomly reducing the number of sampled data of the majority classes, and then re-trains the model on the original dataset. 
In loss-level methods \cite{wang2016training, khan2017cost, lin2017focal, wang2018predicting, zhang2016training, zhang2018image}, rather than modifying the training data distribution, the decision making process is adjusted in a way such that the importance of certain majority classes will be decreased or/and certain pre-defined weights of each class will be take into consideration during the training process. For instance, Lin et al. \cite{lin2017focal} proposes focal loss that reshapes the cross-entropy loss so as to reduce the impact of those easily classified samples and majority classes on the loss during the training process. 
In model-level methods \cite{ren2018multi, garcia2018dynamic, cruz2015meta, zhuang2020cs}, it requires multiple models to be trained on the same dataset and to be fused on the decisions together. The contributions of different models to one final decision are weighted by the model reliability, the model confidence on each sample or both. 
For example, given a pool of classifiers, Ren et al. \cite{ren2018multi} determines the classifier reliability by fuzzy set theory, and combines the decision credibility of each test sample to make the final decisions. 
In summary, most of the current methods aim to tune the training pipeline of CNNs based on the training data class distributions at the data, loss or model levels.

% 3. Existing works mainly drop into following categories:
% (i) Data-level methods for addressing class imbalance include random over-sampling (ROS) \cite{hensman2015impact}, random under-sampling (RUS) \cite{lee2016plankton} and dynamic sampling \cite{pouyanfar2018dynamic}. 
% These thechniques modify the training distribution aim to decrease the level of imbalance, or adjust the sampling rates based on class-wise performance.
% For example, RUS \cite{lee2016plankton} pretrains a model by randomly reduce the sampling number of majority class, and then re-train the model on the original data.
% (ii) Loss-level methods for handling class imbalance do not modify the training data distribution. The decision making process is adjusted in a way that decrease the importance of the majority class, or take a pre-defined weight for each class into consideration \cite{wang2016training, khan2017cost, lin2017focal, wang2018predicting, zhang2016training, zhang2018image}.
% For example, focal loss \cite{lin2017focal} reshapes the cross-entropy loss in order to reduce the impact that easily classified samples and majority class have on the loss. 
% (iii) Model-level methods
% For example, 

However, most of the such existing methods can only work for certain datasets, meaning may not be able to be generalized to different datasets, especially for skin lesion classification. Because the current methods are not designed for addressing one of the fundamental problem of deep learning while dealing with class imbalanced datasets, where the learnt representation of the minority classes is dominated by the majority classes. Hence, it calls for a solution where the good representation of each class should be obtained regardless of data class disproportions. In other words, to solve the problem of enabling good performance of CNNs on class imbalanced datasets, we would like to focus on solving the challenge of learning a good representation that is not affected by the ratio of number of samples. There are a few existing works aiming to enhance the ability of representation learning \cite{misra2020self, wu2018unsupervised, henaff2020data, sermanet2018time, khosla2020supervised}. 
For instance, Misra et al \cite{misra2020self} tries to learn invariant feature representation between the original image and its augmentation by using noise contrastive estimator. 
Khosla et al \cite{khosla2020supervised} leverages the class label to learn clustered representation for downstream tasks. 

% The existing works are trying to give more weights either to the minority class during sampling, or when computing the loss of minority class. 
% However, these solutions may effective sometimes, but fail to address the fundamental problem of deep learning, where the learnt representation of minority class is dominated by the majorities. In another word, good results should be obtained, regardless of class disproportion, if different classes are well represented and come from non-overlapping distributions. Hence, the real challenge is to learn good representations that not affected by the ratio of number of samples.

In this work, we propose to apply and adopt such representation learning techniques in the problem of deep learning for skin lesion classification on class imbalanced datasets. 
We introduce SuperCon, a two-stage training strategy by adopting supervised contrastive loss \cite{khosla2020supervised} and focal loss \cite{lin2017focal} to address class imbalance. 
Supervised contrastive loss learns a clustered feature representation for different classes, and focal loss further enhances the robustness of the classifier. 
To be more specific, during the first stage, we adopt supervised contrastive loss to pull the learnt representations of same class together, and push apart the representations from different classes. 
During the second stage, we employ focal loss to reduce the impact of majority class samples have on the cross-entropy loss. 
The two stages train sequentially, which means the second stage starts after the first stage finished. In that, the classification task in second stage is guided by the learnt feature representation in first stage.
Note that our method neither trying to balance the training data distribution, nor reducing the impact of majority class on loss. We try to address the fundamental problem of deep learning, which is learning a distinguishable representation for different classes, regardless of the class disproportions. 

In the experiments, a comprehensive evaluation using four different backbone networks has been conducted on the current benchmark datasets ISIC challenge 2019 \cite{tschandl2018ham10000, codella2018skin, combalia2019bcn20000} and ISIC challenge 2020 \cite{rotemberg2021patient}. The experimental results show that our SuperCon consistently outperforms the other baseline approaches. Particularly under more class imbalanced dataset, SuperCon performs much better than using focal loss \cite{lin2017focal} during transfer learning by 0.3 in terms of the averaged AUC score among all the backbone networks. 

To summarize, our work makes the following contributions:
\begin{itemize}
    \item We present a novel and effective training strategy for skin image analysis on class imbalance dataset. SuperCon adopts supervised contrastive loss and focal loss to address representation learning and imbalance classification tasks respectively.
    \item To the best of our knowledge, this is the first work that uses supervised contrastive loss to address class imbalance problem, especially for skin lesion classification.
    \item A comprehensive comparison and analysis have been conducted. For the sake of reproducibility and convenience of future studies about skin image analysis on imbalance dataset, we have released our prototype implementation of our proposed SuperCon \footnote{\url{https://github.com/keyu07/SuperCon_ISIC}}. 
\end{itemize}
The rest of the paper is organized as follows: 
Section \ref{relatedwork} presents the related literature review.
Section \ref{methodology} describes our proposed method.
Section \ref{experimentalevaluation} presents the experimental evaluation.
Section \ref{conclusion} presents the conclusion.

\section{Related Work} \label{relatedwork}
\subsection{Skin image analysis in deep learning}
Recent years, extensive research works have applied deep learning in the field of skin image analysis, such as lesion segmentation \cite{li2020transformation,yuan2017automatic, al2018skin, zanddizari2021new} and disease diagnosis \cite{zhang2019attention, mishra2019interpreting, zhuang2020cs, jaworek2019melanoma, barata2019deep, perez2019solo}.
For example, Yuan et al. \cite{yuan2017automatic} presents a end-to-end deep learning framework to automatically segment the dermoscopic images.
Li et al. \cite{li2020transformation} introduces a transformation-consistent strategy in the self-emsenbing model to enhave the regularization effect for pixel-level predictions, such that further improves the skin image segmentation performance.
Zhang et al. \cite{zhang2019attention} proposes an attention neural network model for skin image lesion classification in dermoscopy images and achieves the state-of-the-art performance.
Perez et al. \cite{perez2019solo} investigates the performance of transfer learning among different backbone neural network architectures.
In this paper, we focus on improving the performance of disease diagnosis (classification) under the class imbalance setting. 

\subsection{Class imbalance}
Class imbalance problem refers to that when certain classes (minority classes) contain significantly fewer samples than other certain classes (majority classes). But in many cases, the minority class is the class of interest. For example, ISIC challenge 2020 \cite{rotemberg2021patient} dataset consists of 32,542 benign skin lesion samples and 584 malignant lesion samples, where the malignant type considered as cancerous skin lesion and the class of interest. Conventional deep learning approaches easily over-classify the majority class due to the increased prior probability, resulting in the minority class is misclassified more often. 

In terms of the focus of a deep learning pipeline, existing works to address this problem can be roughly divided into three categories: data-level methods, loss-level methods and model-level methods.
In data-level methods, techniques are usually proposed to adjust the training distribution by tuning the sampling rate of data from different classes \cite{hensman2015impact,lee2016plankton,pouyanfar2018dynamic}. 
For instance, random under-sampling (RUS) \cite{lee2016plankton} pre-trains a model by randomly ignoring the sample from majority class, and re-trains the model on the original dataset.
Pouyanfar et al. \cite{pouyanfar2018dynamic} applies a dynamic sampling method that adjusts sampling rates according to the F1-score of the previous iteration.
In stead of modifying the training data distribution, loss-level methods decrease the importance of certain majority classes during the decision making process, or consider certain pre-defined weights of each class during the training process \cite{wang2016training, khan2017cost, lin2017focal, wang2018predicting, zhang2016training, zhang2018image}.
For instance, Wang et al. \cite{wang2018predicting} modifies the cross-entropy loss to incorporate a pre-defined cost matrix, and forces the model to minimize the misclassification cost during training process. 
Lin et al. \cite{lin2017focal} proposes focal loss by reshaping the cross-entropy loss in order to reduce the impact of those easily classified samples and majority classes on the loss during the training process.
Rather than training a single model, model-level methods normally prepare a pool of models and fuse them to make the final decisions \cite{ren2018multi, garcia2018dynamic, cruz2015meta, zhuang2020cs}. The contributions of different models to one final decision are normally weighted by each model's reliability, the model confidence on each testing sample or both.
For instance, Garcia et al. \cite{garcia2018dynamic} filters the base classifiers by assigning weight to each of them, where the weight is based on the performance of k-nearest neighbors in validation set regarding the current test sample. Finally they fuse the decisions of the filtered classifiers as final decisions.
Ren et al. \cite{ren2018multi} determines the classifier reliability by fuzzy set theory, and combines the decision credibility of each testing sample to make the decisions.

However, the current methods may fail to address the fundamental problem of deep learning while overcoming class imbalanced datasets, where the learnt representation of minority classes is dominated by the majority classes. Hence, it is of great importance to obtain a good representation of each class that regardless of the class disproportions. Some existing works aiming to learn robust feature representation are discussed in next section. 

\subsection{Representation learning} \label{related_representation_learning}
Representation learning in deep learning refers to learn rich and representative patterns from abundance of data. Self-supervised learning and contrastive learning has gained popularity recently because they are able to obtain robust representation via self-defined tasks without using annotating data \cite{misra2020self, wu2018unsupervised, henaff2020data, sermanet2018time, khosla2020supervised, noroozi2016unsupervised}.
For instance, Noroozi et al \cite{noroozi2016unsupervised} aims to re-order the divided image patches, forcing the model to learn the relationship between different parts of an object. 
Wu et al. \cite{misra2020self} tries to learn invariant feature representation between the original image and its augmentation by using noise contrastive estimation.
Khosla et al. \cite{khosla2020supervised} leverages the class label to learn clustered representation for downstream tasks, outperforms the best top-1 accuracy on ResNet200 \cite{he2016deep}.

\section{Methodology} \label{methodology}
To address class imbalance problem, we assume that without balancing the number of samples, a good model should be obtained, regardless of class disproportion, if different classes are well represented and come from non-overlapping distribution. Based on this assumption, we suggest that a good model can learn a feature representation to satisfy the following properties:
% When working with class imbalance, CNN often dominated by the majority class via the conventional training strategy (Transfer learning). For example, when given a dataset with two classes, where 95\% are majority class and the rest are the minority. The network will be accommodated by the majority class and simply predict every sample as the majority. 
% To address this problem, 
(i) The representation should be similar enough within the same class, resulting in a more centralized cluster in the feature space and a easier further classification task. 
(ii) The representation of different classes should be away from each other. Otherwise, overlapping distributions could mislead the classifier to make wrong decisions.

\begin{figure}
\centerline{\includegraphics[width=3.2in]{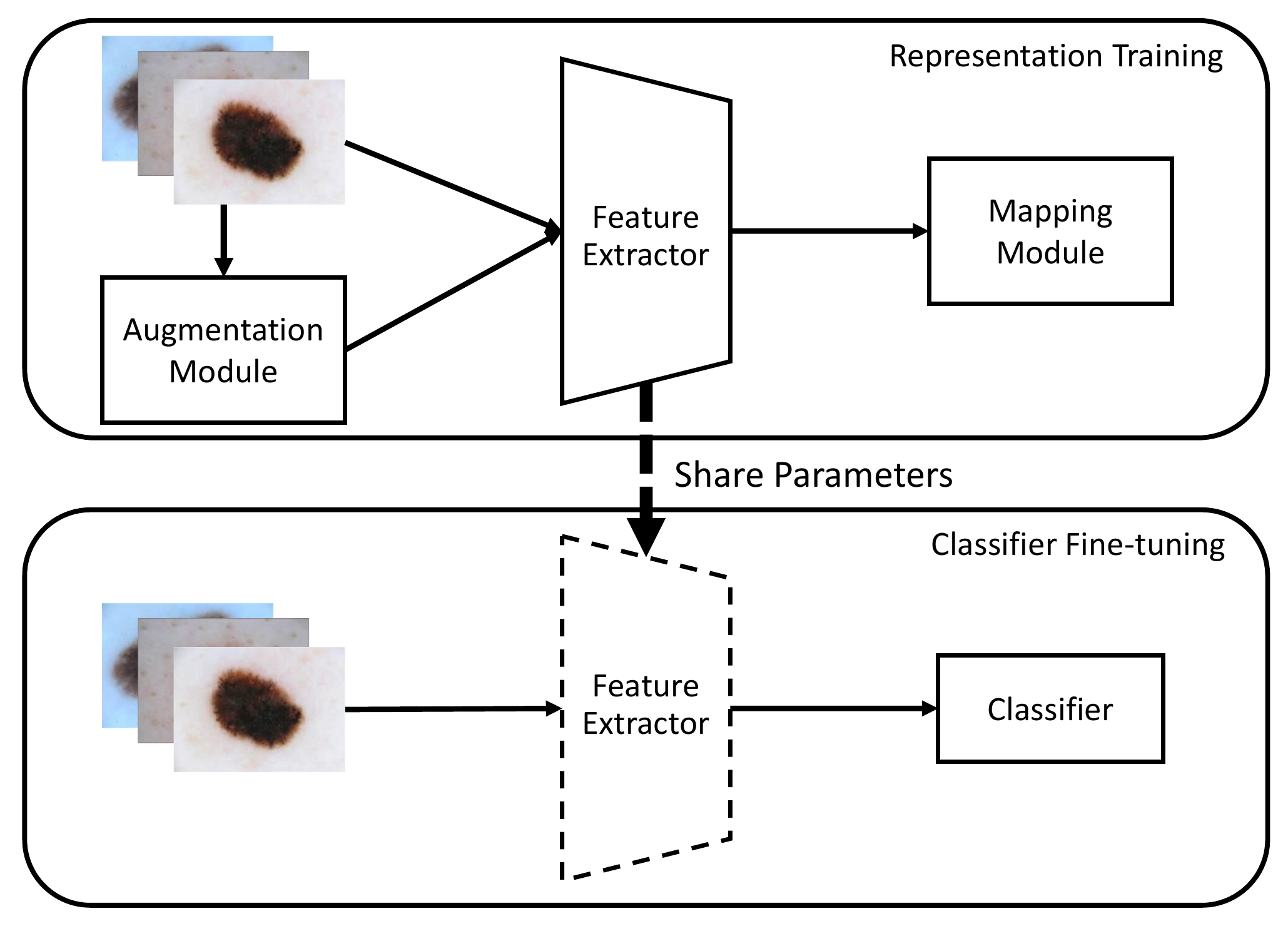}}
\caption{Diagram of our two-stage training. The upper and the bottom branch show the training procedure of representation training and classifier fine-tuning respectively. Feature extractor with dash line means the parameters are fixed.\label{diagram}}
\end{figure}

To overcome the two properties above under the class imbalance setting, we develop SuperCon, a two-stage training strategy to train a CNN model: representation training and classifier fine-tuning.
During the first stage, we employ supervised contrastive loss \cite{khosla2020supervised} to encourage the feature extractor for generating closely aligned feature representation in the same class, and generating distantly representation from different classes. 
In this way, the learnt feature representation is well embedded and guides the classifier to make right decisions.
While during the second stage, we leverage the focal loss \cite{lin2017focal} to reduce the impact that majority class have on the loss.
As illustrated in Figure \ref{diagram}, the upper and the bottom branch show the training procedure of representation training and classifier fine-tuning, respectively. After finishing the representation training, we share and freeze the parameters of well trained feature extractor to further overcome classifier fine-tuning. 
In the rest of this section, we will introduce the two-stage training separately: representation training in section \ref{rep_train} and classifier fine-tuning in section \ref{class_ft}.

\subsection{Representation training} \label{rep_train}
As described above, the goal of this stage is to learn a feature extractor that closely align intra-class feature representation and pull apart the inter-class feature representation.
We consider our representation training including 3 modules (shown in Figure \ref{diagram} upper branch): 
(i) A data augmentation module $A$, which provides different views of the input images. 
(ii) A feature extractor $F_\theta$ with parameter $\theta$, which takes the images as input and encodes them into embedded feature space.
(iii) A mapping module $M$, which maps the embedded feature into a lower dimensional space.
Note that both the feature extractor and the mapping module contain the learnable parameters during training.

For each training sample $x \in X$, we generate one random view $x' \in X'$ via the data augmentation module: 
\begin{equation}
   \textit{$x' = A(x)$}
\end{equation} 
where $x'$ describes the input image in a different view and contains some subset of the information in the original image. The view could be different rotation angle, color distortion, etc. We consider that the various views provide additional noise to the feature extractor and enhance its generalization ability. The details of the augmentation methods used in our experiments will be introduced in section \ref{exp_setting}. 

Then the feature extractor encodes the original input and its augmentation to a feature representation vector $rep$, respectively:
\begin{equation}
   \textit{$rep,~rep' = F_{\theta}(x),~F_{\theta}(x')$}
\end{equation}

For each mini-batch, it includes the original inputs and the augmented inputs. That turns the mini-batch into a multiviewed-batch. 

To save the expense of computing supervised contrastive loss, we map both feature representations to a lower dimension using the mapping module:

\begin{equation}
   \textit{$z,~z' = M(rep),~ M(rep')$  }
\end{equation}
where $z$ and $z'$ represent the lower dimension of feature representation of original input and its view. After this, the supervised contrastive loss is applied on $z$ and $z'$ as following: 

\begin{equation}
   \textit{$\mathcal{L}_{SuperCon} = \sum_{x,x' \in X,X'}\frac{-1}{|P|}\sum_{p\in P} log \frac{exp(z_i \centerdot z_p / \tau)}{\sum_{n \in N}exp(z_i \centerdot z_n / \tau)}$}  \label{supconloss}
\end{equation}
where the index $i$ is called the anchor, the $p$ is the index of positive samples $P$, the $n$ is the index of negative samples $N$, and $|P|$ is the number of positive samples in the multiviewed-batch. The positives and negatives are the samples have same class labels as the anchor and have different labels from the anchor, respectively. Note that both $z$ and $z'$ could be in the anchor, positives or negatives, because we only consider the class label in this loss. The dot product computes the cosine similarity of the low dimension representation between the anchor and the positive or negative. $\tau$ is a scalar temperature parameter.
Recall that our objective is to closely align intra-class feature representation and distantly apart the inter-class feature representation. It means in feature space, the cosine similarity of intra-class should be close to 1, and the inter-class should be close to 0. Hence, by minimizing the supervised contrastive loss, we enforce the numerator close to 1 and the denominator close to 0, and achieve our objective. 

To summarize, we update the parameters of the feature extractor as follows:
\begin{equation}
   \textit{$\theta_{m+1} \gets \theta_m - \lambda \cdot \nabla(\mathcal{L}_{SuperCon}) $}  \label{featureextractor_update}
\end{equation}
where the $m$ and $\lambda$ indicate the index of iteration and learning rate respectively.

\subsection{Classifier fine-tuning} \label{class_ft}
Since the feature extractor is well trained in previous stage, we apply classifier fine-tuning to address the final label prediction task. A common issue of class-imbalanced data training is that the easily classified majority class samples dominate the gradient, but the minority class has not been paid enough attention. Finally the model is over-confident on the majority but failed on minority. This issue is mitigated by using focal loss \cite{lin2017focal}, which reduces the impact that majority class have on the loss.

To train the classifier, the classification model is composed of the feature extractor $F_\theta$ from representation training and a trainable classifier $C_\varphi$ with learnable parameters $\varphi$. Since the feature extractor has learnt clustered feature representation, we freeze the parameters of feature extractor $\theta$ and only fine-tune the classifier in this stage. By using the focal loss, which already shown a performance improvement on class imbalance problem \cite{lin2017focal}. It reshapes the cross-entropy loss and assign different weight to majority and minority classes. The loss function defined as follows:

\begin{equation}
   \textit{$ \mathcal{L}_{Focal}(Pro_t) = -\alpha_t(1-Pro_t)^{\gamma}log(Pro_t)$}  \label{focalloss}
\end{equation}
where $\alpha_t$ is a class-wise weight that is used to increase the importance of the minority class. While $\gamma$ is a sample-wise weight that is used to reduce the propagation impact from well-classified samples. $t$ and $Pro$ indicate the index of the sample and the related predicted probability.

The classifier parameter updated as follows:
\begin{equation}
   \textit{$\varphi_{m+1} \gets \varphi_m - \beta \cdot \nabla(\mathcal{L}_{Focal}) $}  \label{classifier_update}
\end{equation}
where the $\beta$ is the classification learning rate.

\section{Experimental Evaluation} \label{experimentalevaluation}
We conduct our experiments on two benchmark datasets, and use ResNet backbone networks \cite{he2016deep} to evaluate the performance of our proposed SuperCon. 
Comprehensive comparisons have been made among SuperCon and other baseline approaches. 
The presented results are shown that SuperCon significantly improves other approaches and achieves state-of-the-art performance on these two datasets.

\subsection{Experimental datasets}
In our experiment, we utilize the well known ISIC Challenge 2020 dataset \cite{rotemberg2021patient}, and some additional data from ISIC Challenge 2019 \cite{tschandl2018ham10000, codella2018skin, combalia2019bcn20000}. 
The ground truth of both original testing data are not given, so we only employ the original training data without meta-data in our evaluation. 
ISIC 2020 dataset consists of 33,126 dermoscopic images, where 584 are confirmed malignant skin lesions and 32,542 are benign lesions.
We split the entire 33,126 images into training set (including 26,003 benign and 467 malignant skin lesion cases) and testing set (6,509 benign and 117 malignant cases). 
In order to mitigate the imbalance ratio, we add the 4,522 melanoma images from ISIC 2019 to our training set of ISIC 2020, to expand the number of samples in malignant case.
Moreover, the evaluation analysis on only ISIC 2020 also conducted in section \ref{exp_2020only}.

\subsection{Experimental settings} \label{exp_setting}

During our two-stage training process, 
we first employ representation training to train a feature extractor that good at capturing the representations of the training images. 
Then, we freeze the parameters of feature extractor and fine-tune the classifier on training set using Stochastic Gradient Descent (SGD) optimizer for 10 epochs.
The loss converges quickly in second stage due to the well learnt feature extractor from the first stage, so we do not need many number of epochs to train the classifier.
Both stages use batch size 128 to ensure at least one anchor in the mini-batch.
For both stage training, we randomly apply color distortion, grayscale, horizontal flip and Gaussian blur to provide the multi-views of the input images.
The mapping module $m$ takes the output of the feature extractor and reduce its dimension to 128.
We follow \cite{khosla2020supervised} to choose the hyperparameters for the first stage: $\tau$ = 0.1 and learning rate $\lambda$ = 0.01.
During the second stage, since the classifier only needs to fine-tune, so we simply use small learning $\beta$ = 5e-4, $\alpha$ = 0.25 and $\gamma$ = 5 for focal loss. Note that the final results are not sensitive to the change of $\alpha$ and $\gamma$. 
We adopt ResNets pre-trained on ImageNet \cite{deng2009imagenet} as our backbone networks to evaluate our proposed approach: ResNet18, ResNet50, ResNet101 and ResNet152 \cite{he2016deep}.
Model-agnostic can be achieved by simply changing the backbone network architectures without additional implementation. 
All of our experiments are implemented using PyTorch, on a server with a RTX 3090 24 GB GPU.

\subsection{Effectiveness analysis}
To demonstrate the effectiveness of SuperCon, a comparison among other baseline approaches and ours has been conducted. These approaches are evaluated on the ResNets \cite{he2016deep} we mentioned in section \ref{exp_setting}:
\begin{itemize}
    \item Vanilla is the baseline that trains the given dataset on the pre-trained CNN network with cross-entropy loss. 
    \item Focal-Loss \cite{lin2017focal} has the same setting as the Vanilla but uses the focal loss.
    \item ROS \cite{hensman2015impact} Random Over-Sampling trains the given dataset by over-sampling the minority class, which is the malignant skin lesion.
    \item RUS \cite{lee2016plankton} Random Under-Sampling randomly pre-trains a model by randomly reducing the number of sampled data in majority class, and then re-train the model on the original dataset.
    \item SuperCon-CE is the approach that employs our two-stage training strategy, but uses cross-entropy loss during the second stage (classifier fine-tuning).
    \item SuperCon is our full two-stage training strategy with focal loss.
\end{itemize}

Given a baseline approach, we evaluate its effectiveness in terms of the Macro F1-score and AUC score.
\begin{equation}
   \textit{$ Precision=\frac{TP}{TP + FP} $}  \label{precision}
\end{equation}
\begin{equation}
   \textit{$ Recall=\frac{TP}{TP + FN}$}  \label{recall}
\end{equation}
\begin{equation}
   \textit{$ Micro~F1=2\times \frac{Precision\times Recall}{Precision + Recall}$}  \label{MicroF1}
\end{equation}
\begin{equation}
   \textit{$ Macro~F1=\frac{1}{|C|} \sum_{C_i}^{C}F1(C_i)$}  \label{MacroF1}
\end{equation}

where in equation \ref{precision}, TP and FP are True Positives and False Positives, respectively. FN in equation \ref{recall} is False Negatives. In equation \ref{MacroF1}, $|C|$ represents the total number of classes in the dataset, and $F1(C_i)$ indicates the F1-score for each single class.
When working with imbalanced dataset, using Micro F1-score can be misleading. Because the Micro F1-score gives equal importance to each sample, and the majority classes will drive a huge portion of this score. While Macro F1-score gives equal importance to each class, and the majority and minority classes will drive equal portion of this score. Thus, in the experimental evaluation, we only use Macro F1-score and present F1-score for short.

The results are shown in Table \ref{tab1}, we can observe that:
(i) With deeper backbone network architecture, both F1-score and AUC score increase on all the approaches . 
(ii) Vanilla always has reasonable performance in terms of AUCs. But the F1-scores are very low. 
(iii) Focal-Loss and RUS have slightly improvement on the Vanilla in both scores.
(iv) ROS always has the worst performance, even worse than the baseline Vanilla. 
(v) Our SuperCon consistently outperforms the others in terms of AUC score and only slightly lower than that without using focal loss on ResNet101.
(vi) If we compare the last two columns (SuperCon-CE and SuperCon) with Focal-Loss. We can see that the performance improvement is mainly coming from the representation training.
(vii) On the overall average using the 4 different backbone networks, SuperCon-CE and SuperCon outperform the other approaches by at least 0.1 on F1-score and over 0.04 on AUC score, respectively.

\begin{figure}[!h]
\centering
\subfloat[]{\label{1}\includegraphics[width=0.45\linewidth]{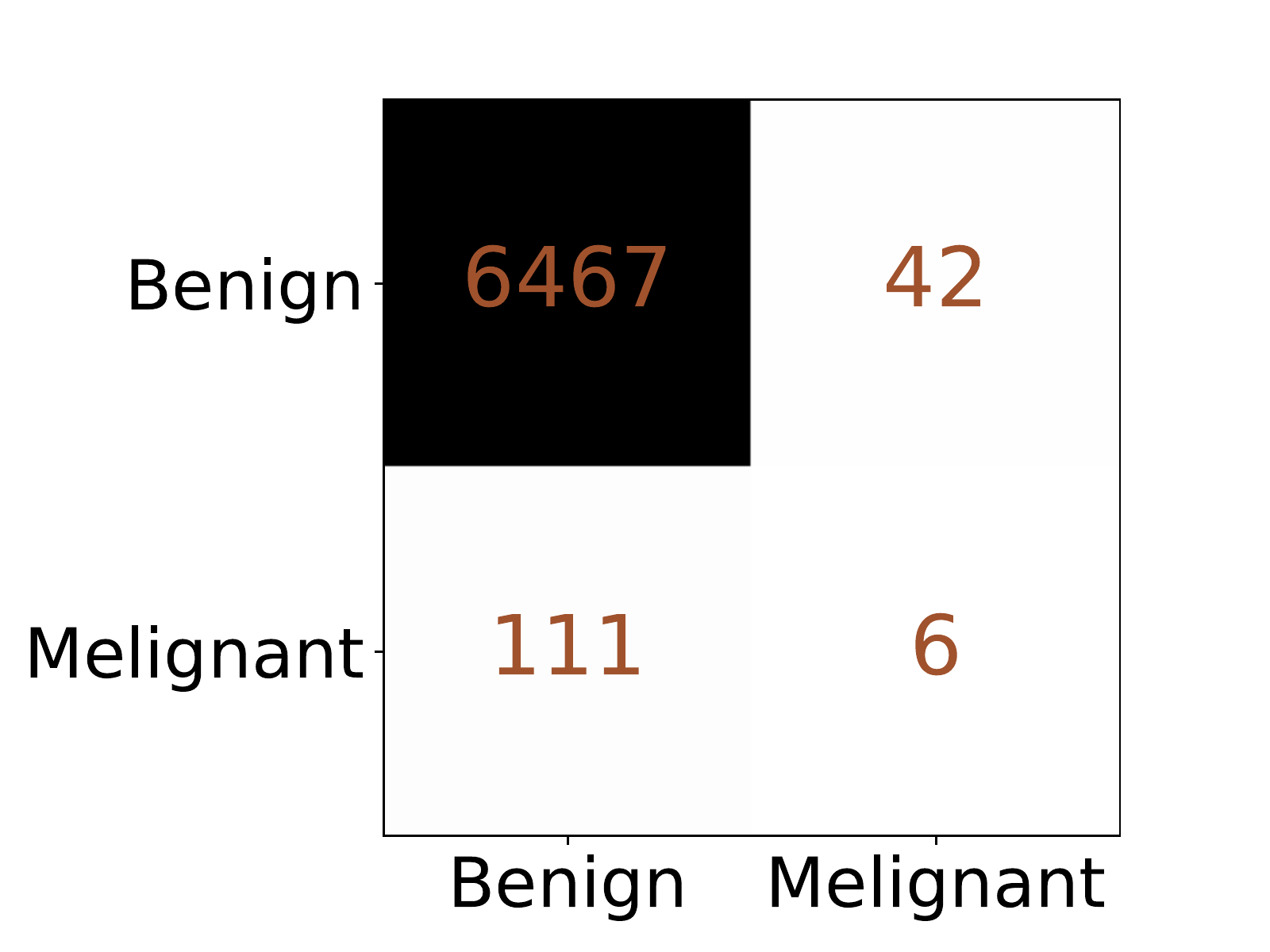}}\hfil
\subfloat[]{\label{2}\includegraphics[width=0.45\linewidth]{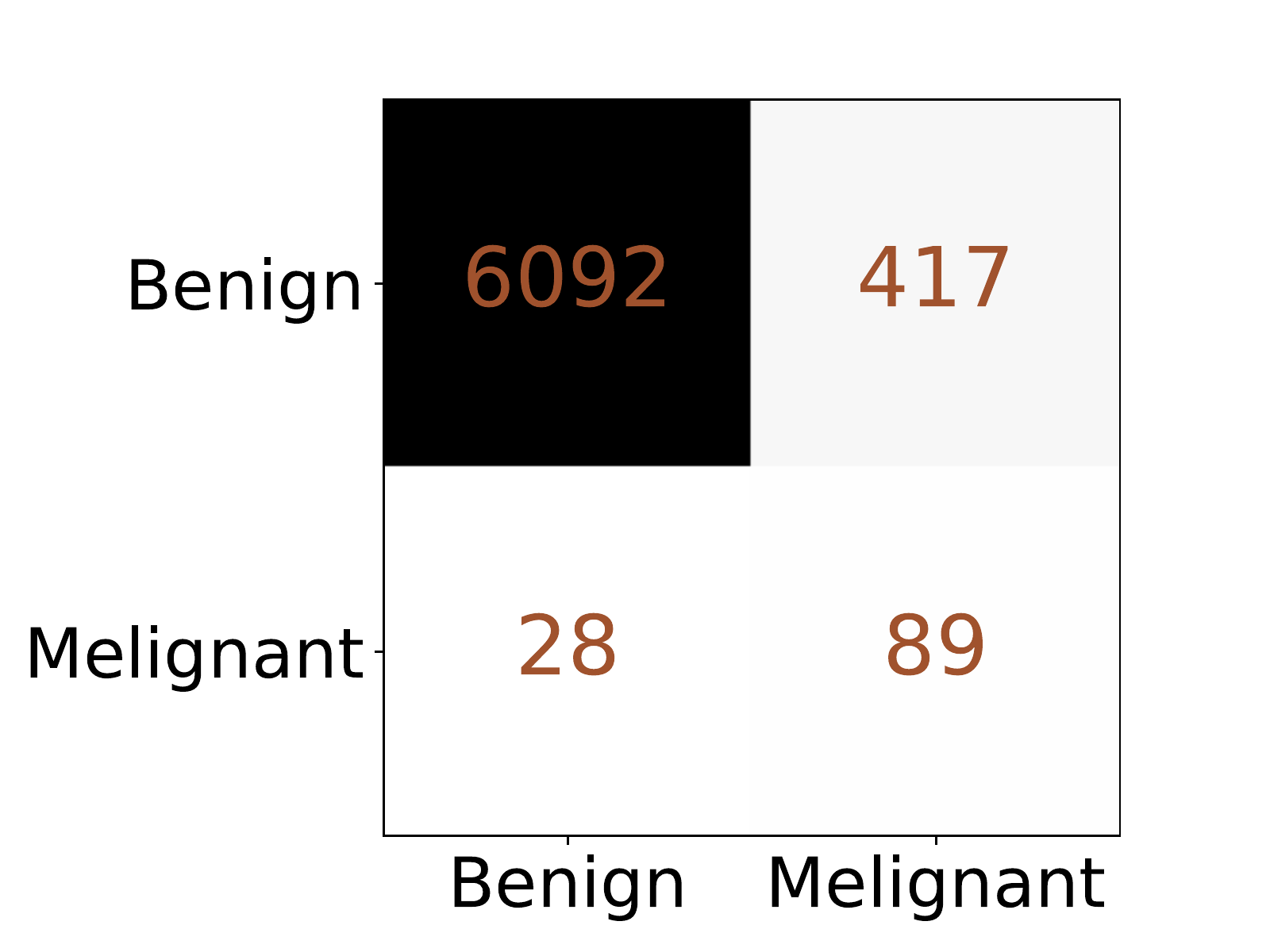}}\hfil\\
\subfloat[]{\label{3}\includegraphics[width=0.45\linewidth]{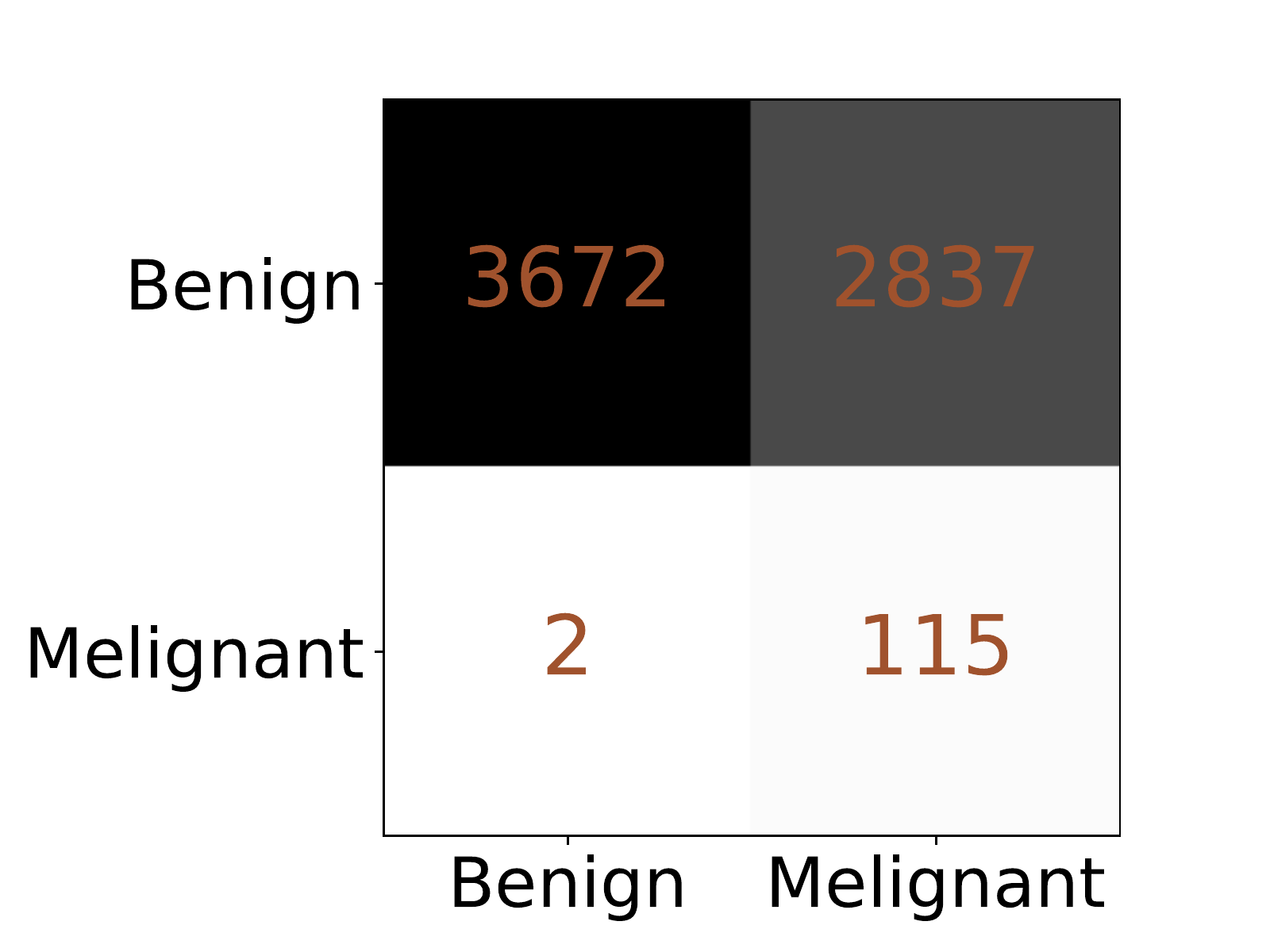}}\hfil
\subfloat[]{\label{4}\includegraphics[width=0.45\linewidth]{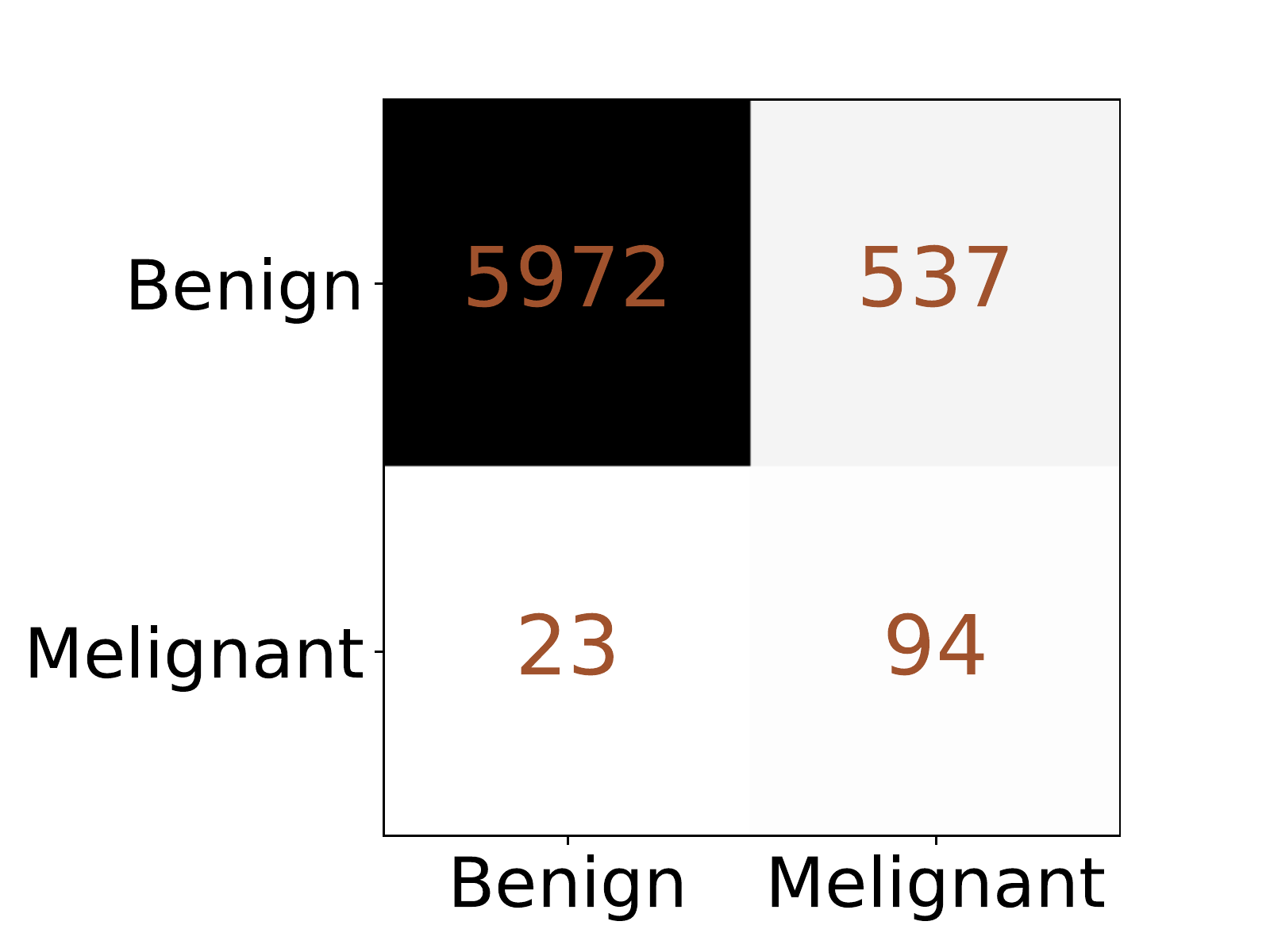}}\hfil\\
\subfloat[]{\label{3}\includegraphics[width=0.45\linewidth]{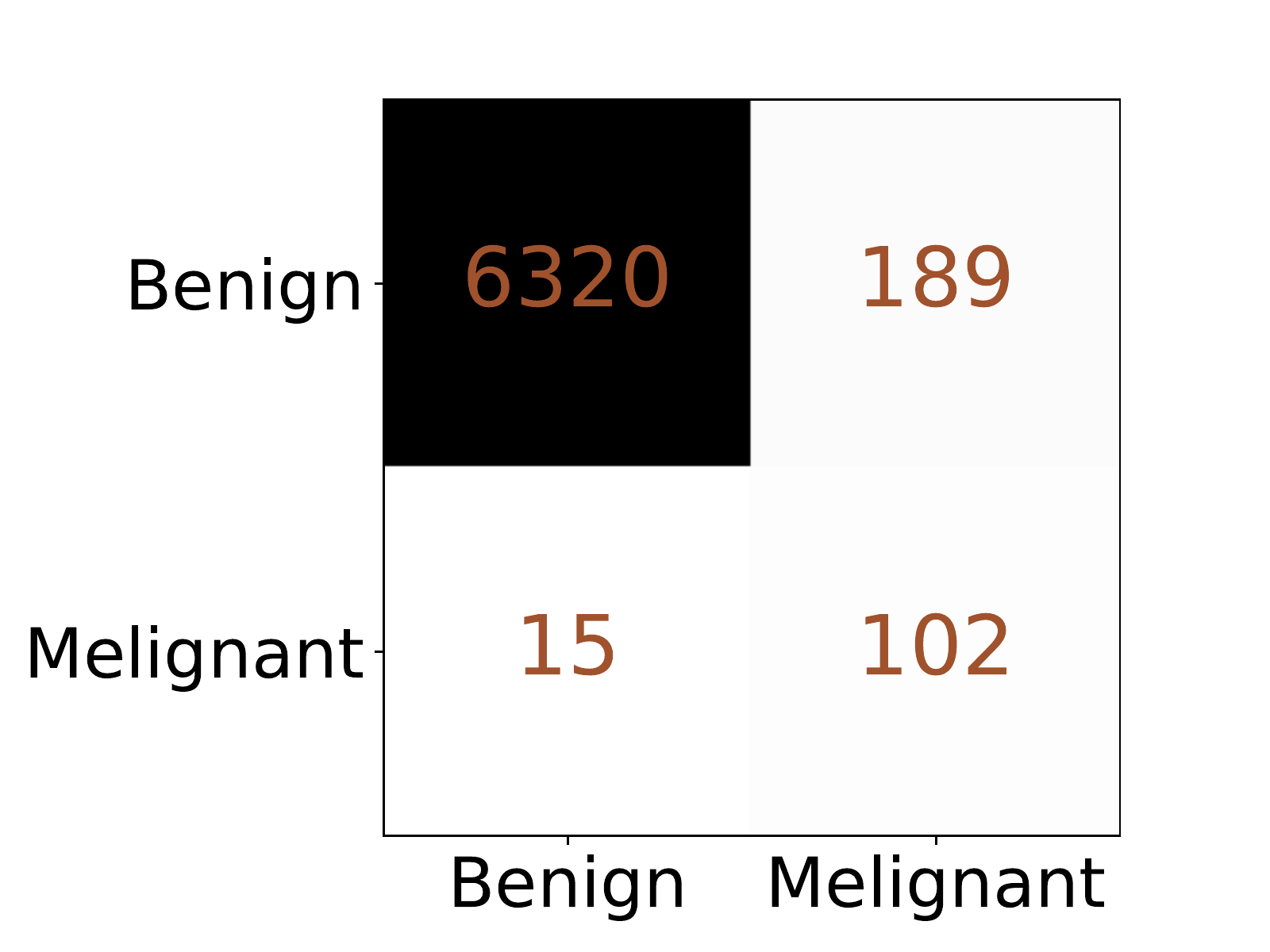}}\hfil
\subfloat[]{\label{4}\includegraphics[width=0.45\linewidth]{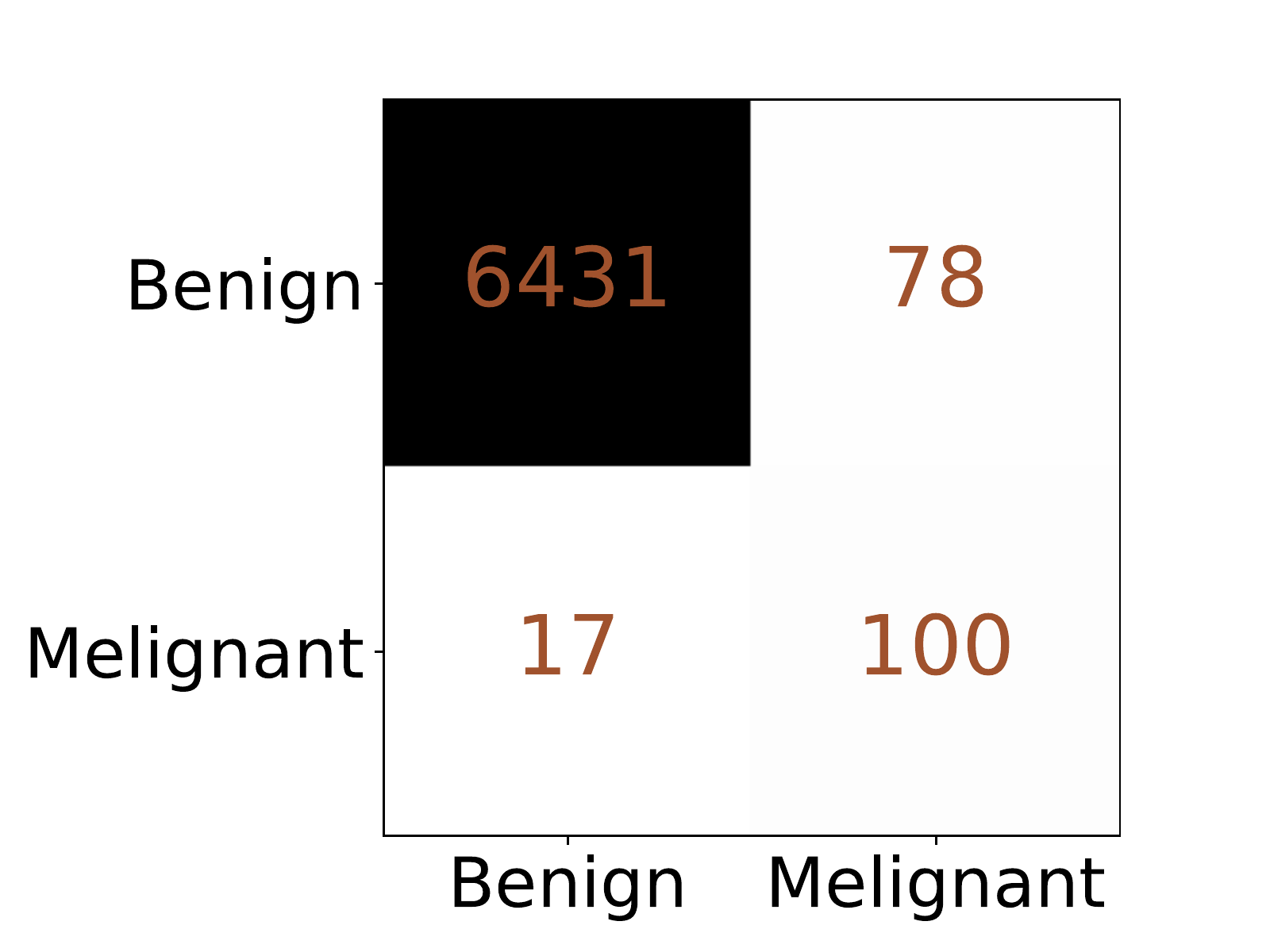}}\hfil\\
\caption{Confusion matrices of four approaches using backbone network ResNet152 on dataset ISIC 2019+2020. (a) Vanilla; (b) Focal-Loss; (c) ROS; (d) RUS; (e) SuerCon-CE; (f) SuperCon.}\label{confusionmatrix1}
\end{figure}

To better demonstrate the effectiveness of SuperCon, confusion matrices of the six approaches using ResNet152 are plotted in Figure \ref{confusionmatrix1}, the x-axis and y-axis indicate the predicted and true label.
We can see that:
(i) In Vanilla, the minority class is dominated by the majority class, where 111 out of 117 samples are classified as majority class.
(ii) Compare to Vanilla, Focal-Loss leads a better classification performance on True Negative, where 89 samples in melanoma are correctly classified. But it also has a higher False Negative, where 417 samples in majority class are misclassified.
(iii) In Figure \ref{confusionmatrix1}(c). ROS shows a very high False Negative, where over 40\% majorities are misclassified. Because, by design, ROS uniformly learns the minority and majority classes, resulting in overperformed on minority and underperformed on majority class.
(iv) RUS also tries to balanced the distribution by randomly ignore the majority samples, but it fine-tunes on the original data distribution afterward. Thus it has much lower False Negative than ROS.
(v) SuperCon-CE and SuperCon have significantly improvement in terms of the number of correctly classified samples on both majority and minority class.

\begin{table}
  \renewcommand\arraystretch{1.1}
  \caption{Experimental results on the ISIC 2020 with additional data from ISIC 2019, using different ResNet backbone networks. The results are the average over 3 repretitions. Best performance in bold.}
  \label{tab1}
  \scalebox{0.67}{
  \begin{tabular}{ccccccc}
    \toprule
    \textbf{ISIC} & \multirow{2}{*}{\textbf{Vanilla}} & \multirow{2}{*}{\textbf{Focal-Loss}} 
    & \multirow{2}{*}{\textbf{ROS}} & \multirow{2}{*}{\textbf{RUS}}
    & \textbf{SuperCon} & \multirow{2}{*}{\textbf{SuperCon}}\\
     \textbf{2019+2020} & & & &  & \textbf{-CE} & \\
    \midrule
    \multicolumn{7}{c}{\textbf{ResNet18}} \\
    \midrule
     F1-score   & 0.51 & 0.57 & 0.35 & 0.56 & 0.56  & \textbf{0.58} \\
     AUC score  & 0.742 & 0.760 & 0.725 & 0.794 & 0.773  & \textbf{0.778}  \\
     \midrule
     \multicolumn{7}{c}{\textbf{ResNet50}} \\
     \midrule
     F1-score   & 0.51 & 0.58 & 0.36 & 0.58 &\textbf{0.67}  & \textbf{0.67}\\
     AUC score  & 0.791 & 0.833 & 0.744 & 0.826 & 0.888  & \textbf{0.892}  \\
    \midrule
     \multicolumn{7}{c}{\textbf{ResNet101}} \\
     \midrule
     F1-score  & 0.56 & 0.64 & 0.38  & 0.62 & \textbf{0.75}  & 0.68 \\
     AUC score  & 0.817 & 0.845 & 0.760 & 0.854 & \textbf{0.923}  &  0.919 \\
     \midrule
     \multicolumn{7}{c}{\textbf{ResNet152}} \\
     \midrule
     F1-score   & 0.53 & 0.63 & 0.40 & 0.60 & 0.74  & \textbf{0.84} \\
     AUC score  & 0.834 & 0.848 & 0.774 & 0.860 & \textbf{0.921} & \textbf{0.921} \\
     \midrule
     \multicolumn{7}{c}{\textbf{Average}} \\
     \midrule
     F1-score  & 0.53 & 0.61 & 0.37 & 0.59 & 0.68  & \textbf{0.69} \\
     AUC score & 0.796 & 0.822 & 0.751 & 0.834 & 0.876 & \textbf{0.878} \\
    \bottomrule
\end{tabular}}
\end{table}

\begin{table}
  \renewcommand\arraystretch{1.1}
  \caption{Experimental results only on the ISIC 2020, using different ResNet backbone networks. The results are the average over 3 repretitions. Best performance in bold.}
  \label{tab2}
  \scalebox{0.75}{
  \begin{tabular}{cccccc}
    \toprule
    \multirow{2}{*}{\textbf{ISIC 2020}} & \multirow{2}{*}{\textbf{Vanilla}} & \multirow{2}{*}{\textbf{Focal-Loss}} 
    & \multirow{2}{*}{\textbf{ROS}} & \textbf{SuperCon} & \multirow{2}{*}{\textbf{SuperCon}}\\
      & & & & \textbf{-CE} & \\
    \midrule
    \multicolumn{6}{c}{\textbf{ResNet18}} \\
    \midrule
     F1-score & 0.50 & 0.53 & 0.43 & \textbf{0.79} & \textbf{0.79}   \\
     AUC score & 0.500 & 0.552 & \textbf{0.804} & 0.709 & 0.726    \\
     \midrule
     \multicolumn{6}{c}{\textbf{ResNet50}} \\
     \midrule
     F1-score   & 0.50 & 0.55 & 0.43 & 0.86  & \textbf{0.89} \\
     AUC score  & 0.500 & 0.542 & 0.809 & 0.865  & \textbf{0.876}  \\
    \midrule
     \multicolumn{6}{c}{\textbf{ResNet101}} \\
     \midrule
     F1-score  & 0.50 & 0.57 & 0.43 & 0.93 & \textbf{0.94}  \\
     AUC score  & 0.500 & 0.558 & 0.810 &  0.897  & \textbf{0.901}  \\
     \midrule
     \multicolumn{6}{c}{\textbf{ResNet152}} \\
     \midrule
     F1-score   & 0.50 & 0.55 & 0.44 & 0.93 & \textbf{0.94}  \\
     AUC score  & 0.500 & 0.554 & 0.821 & 0.903 &  \textbf{0.906} \\
     \midrule
     \multicolumn{6}{c}{\textbf{Average}} \\
     \midrule
     F1-score & 0.50 & 0.55 & 0.43 & 0.88  & \textbf{0.89} \\
     AUC score & 0.500 & 0.552 & 0.811 & 0.843 & \textbf{0.852} \\
    \bottomrule
\end{tabular}}
\end{table}

\begin{figure*}
\centering
\subfloat[]{\label{1}\includegraphics[width=0.2\linewidth]{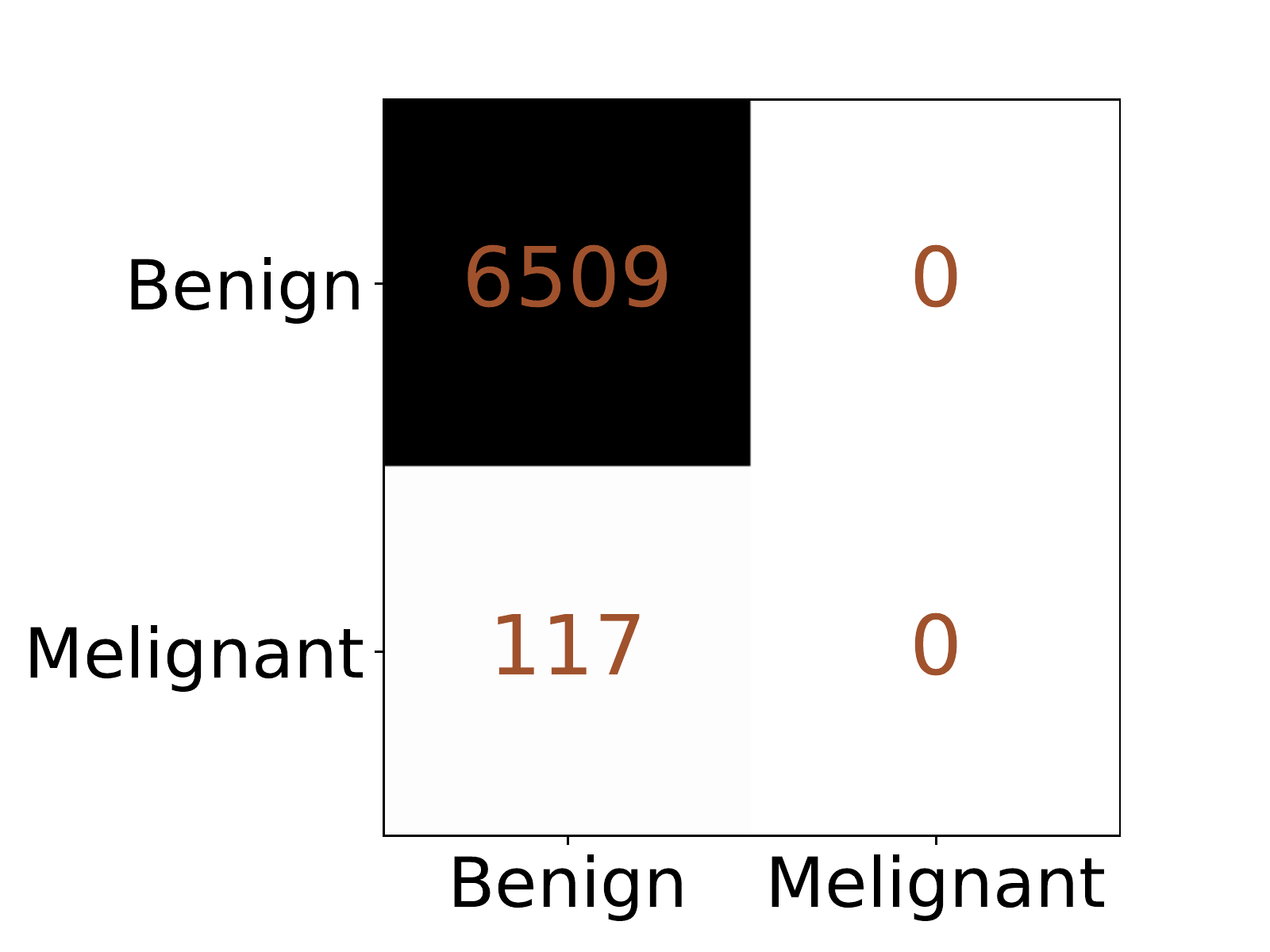}}\hfil
\subfloat[]{\label{2}\includegraphics[width=0.2\linewidth]{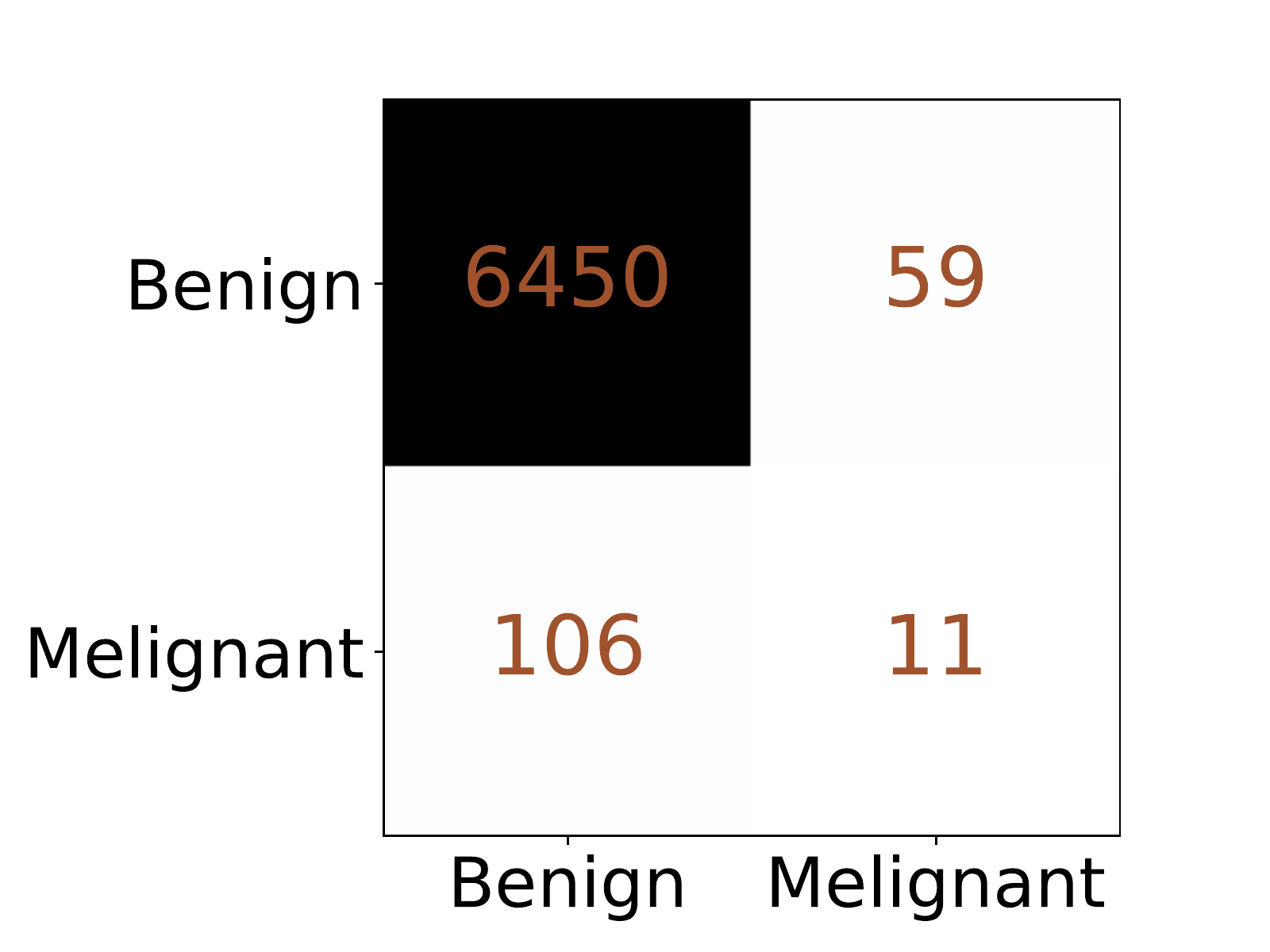}}\hfil
\subfloat[]{\label{3}\includegraphics[width=0.2\linewidth]{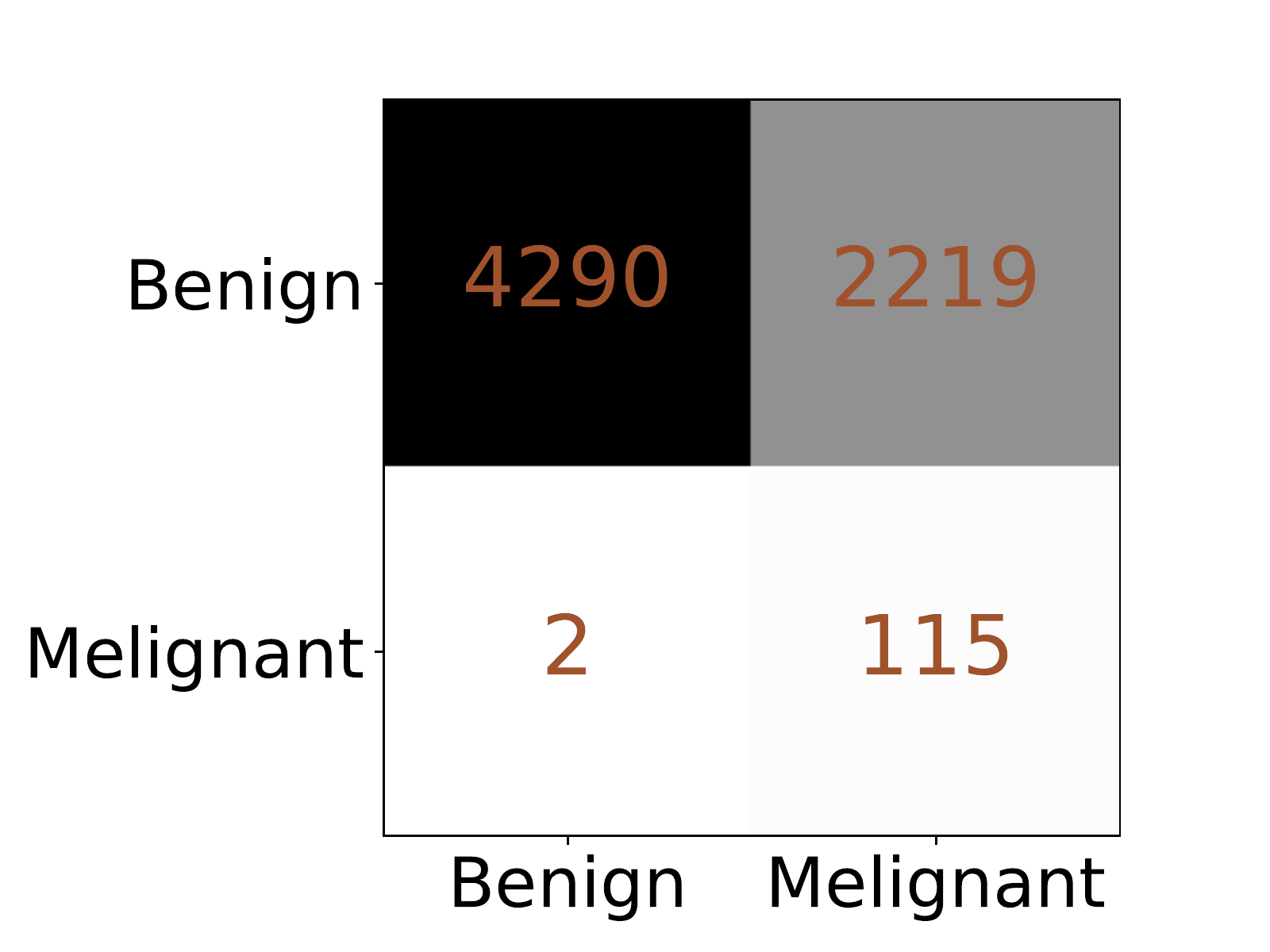}}\hfil
\subfloat[]{\label{4}\includegraphics[width=0.2\linewidth]{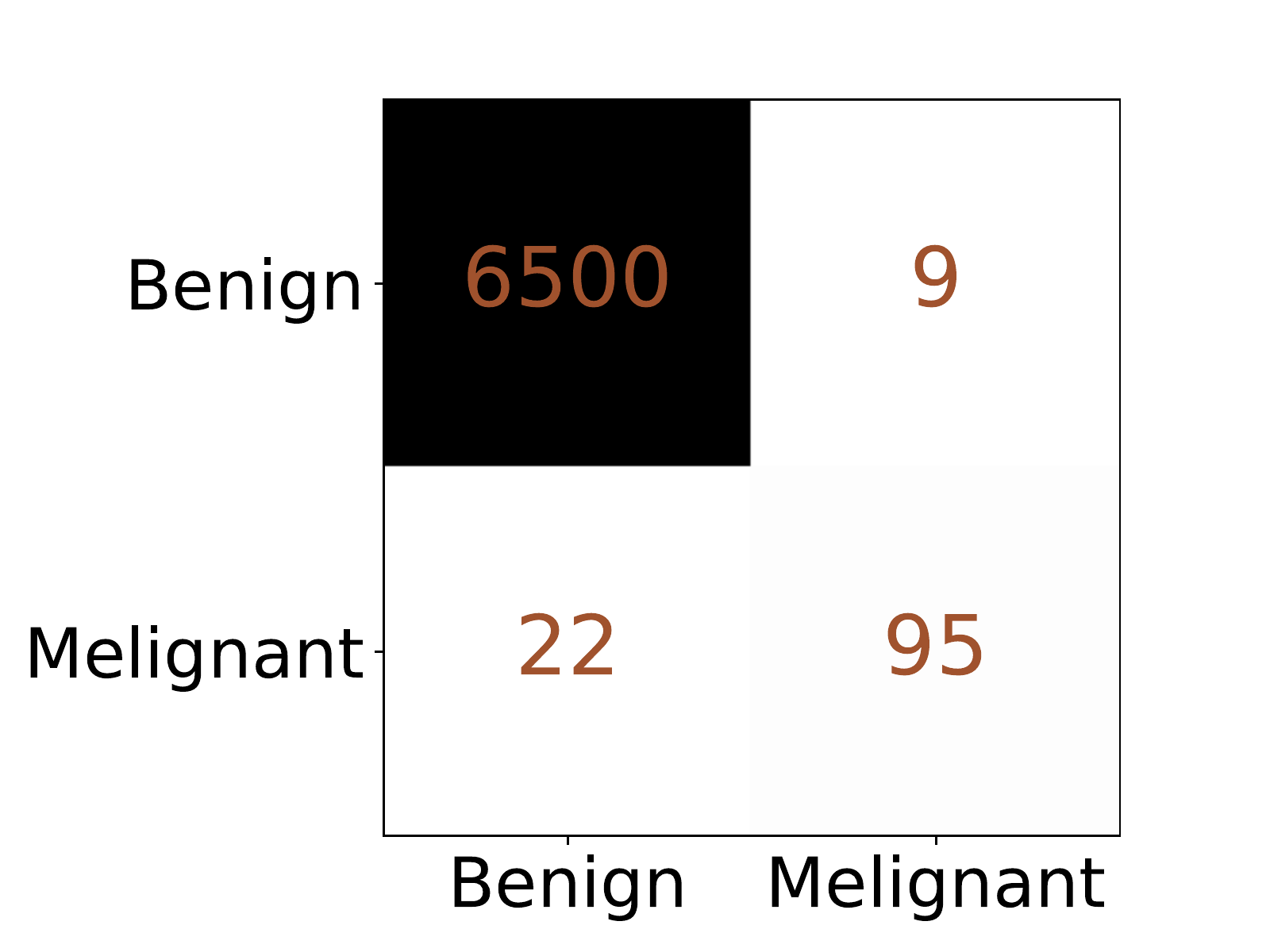}}
\subfloat[]{\label{5}\includegraphics[width=0.2\linewidth]{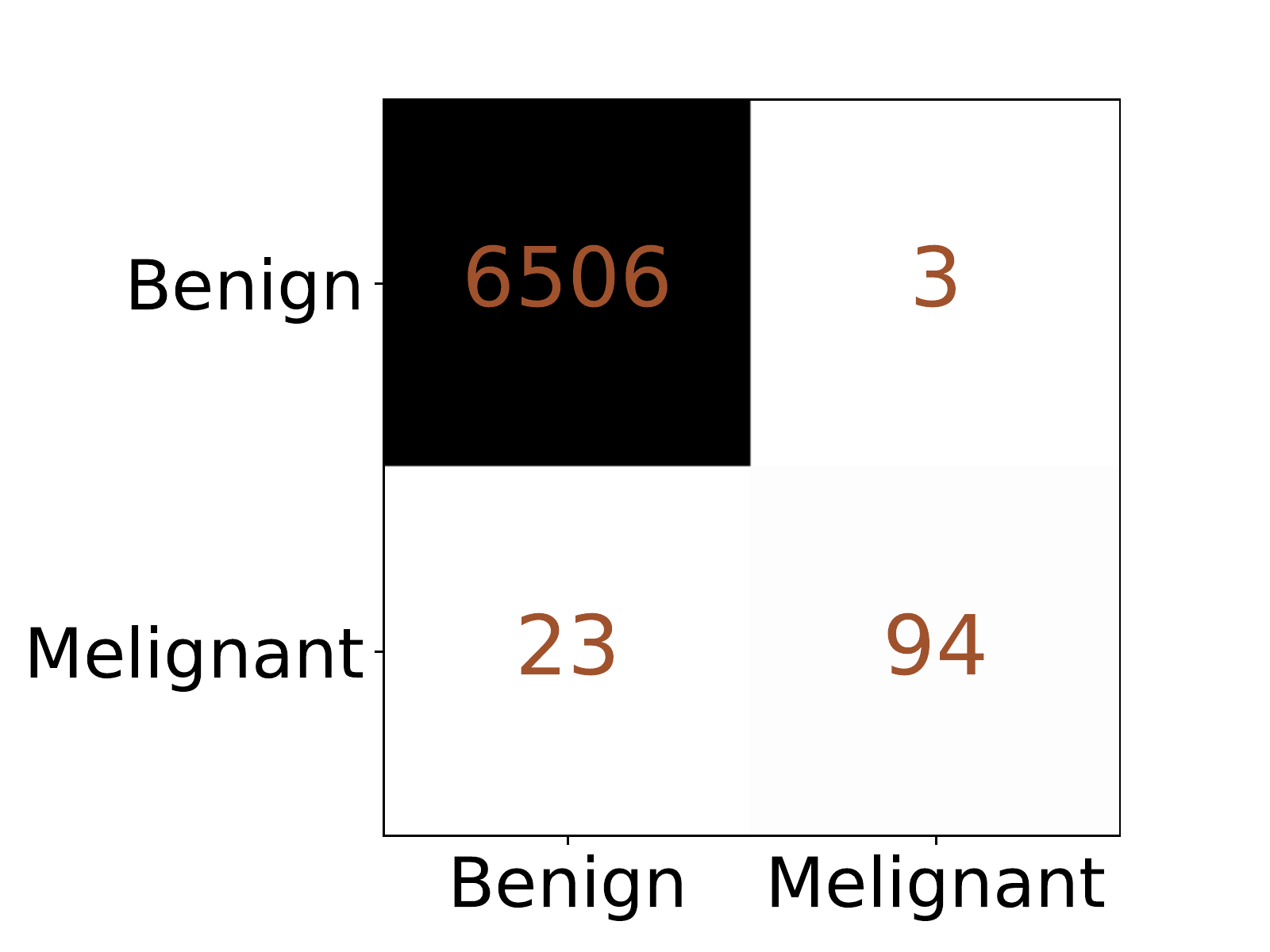}}
\caption{Confusion matrix of each approach using backbone network ResNet152 on dataset ISIC 2020. (a) Vanilla; (b) Focal-Loss; (c) ROS; (d) SuperCon-CE; (e) SuperCon. }\label{confusionmatrix2}
\end{figure*}
\begin{figure*}
\centering
\subfloat[]{\label{1}\includegraphics[width=0.33\linewidth]{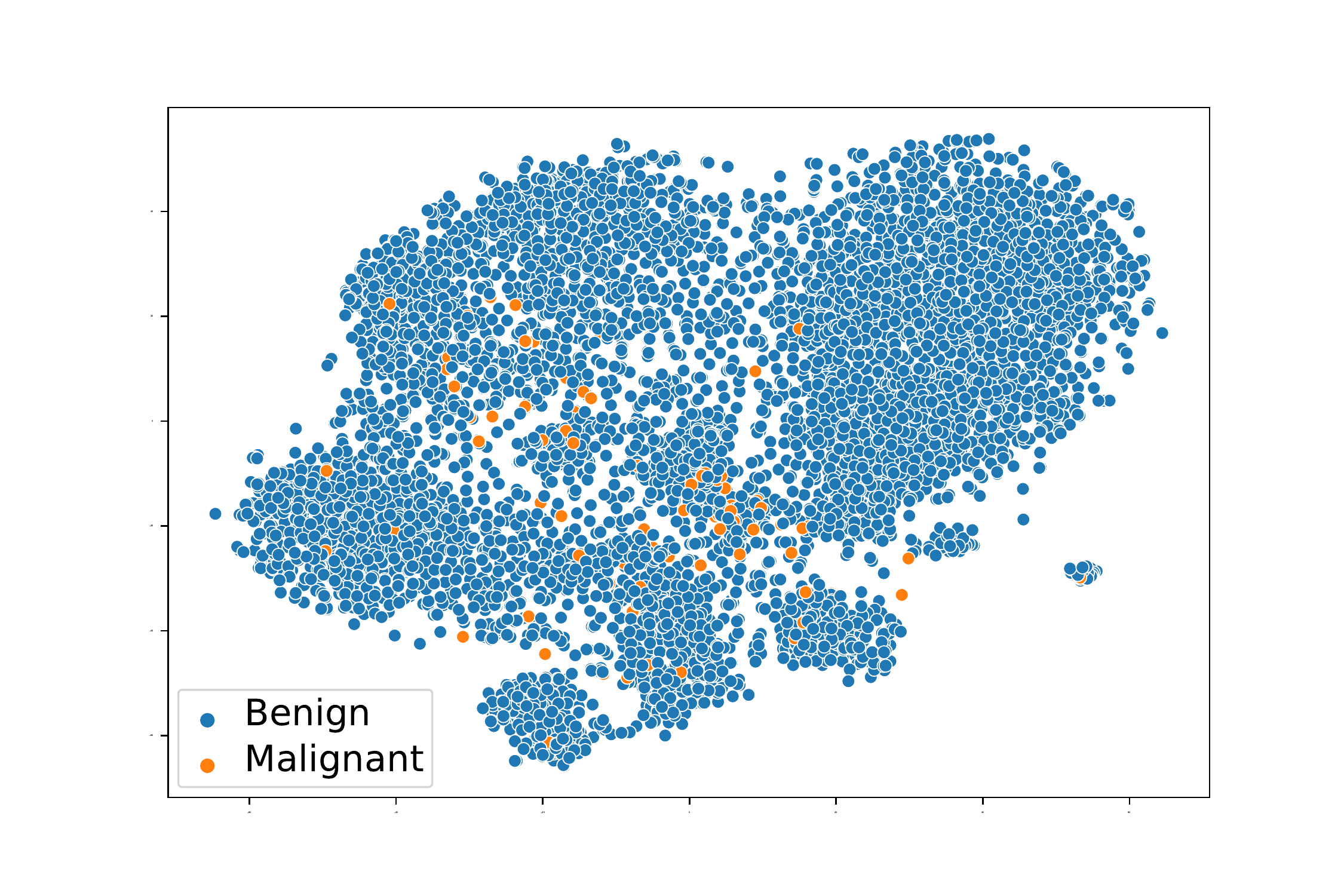}}\hfil
\subfloat[]{\label{2}\includegraphics[width=0.33\linewidth]{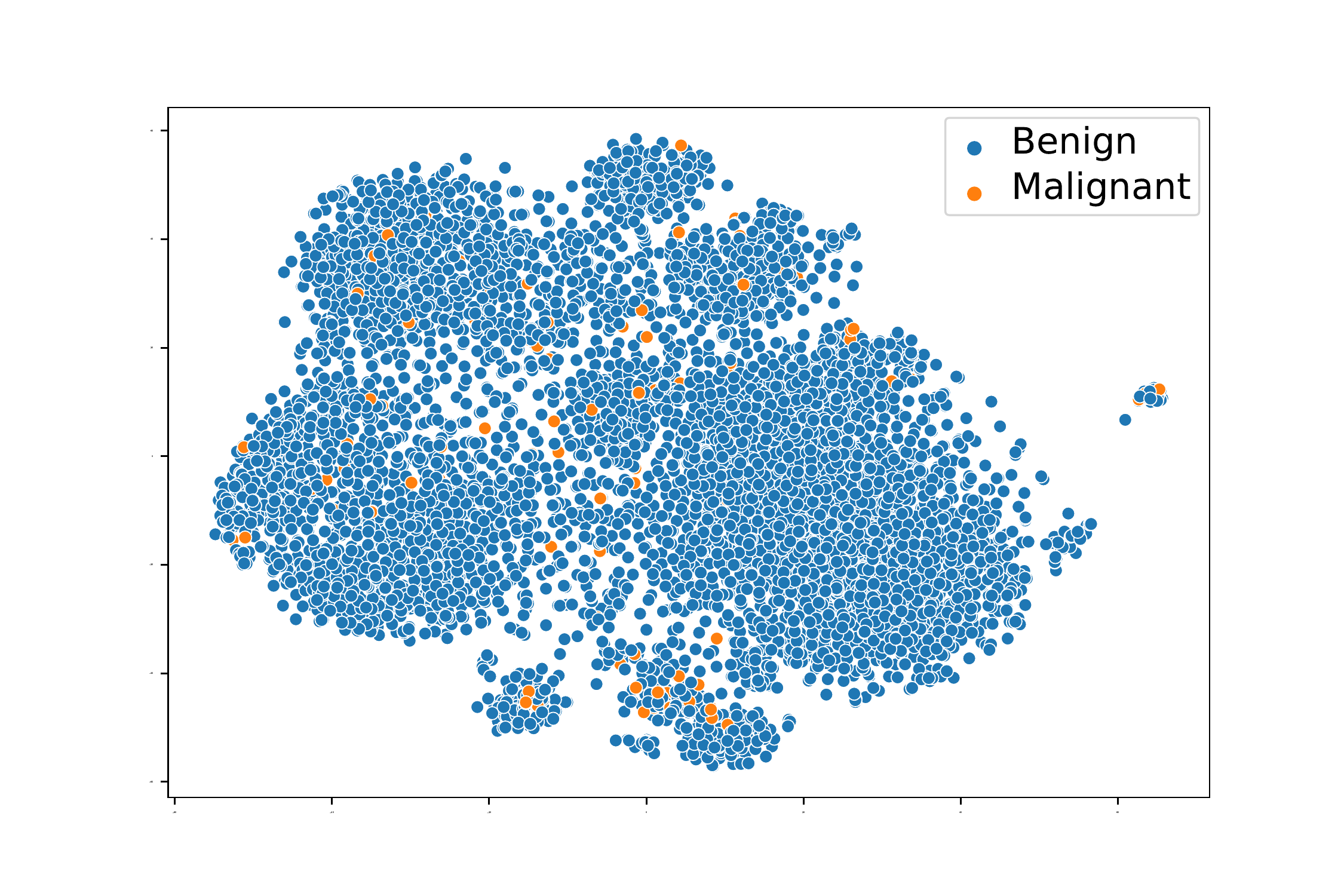}}\hfil
\subfloat[]{\label{3}\includegraphics[width=0.33\linewidth]{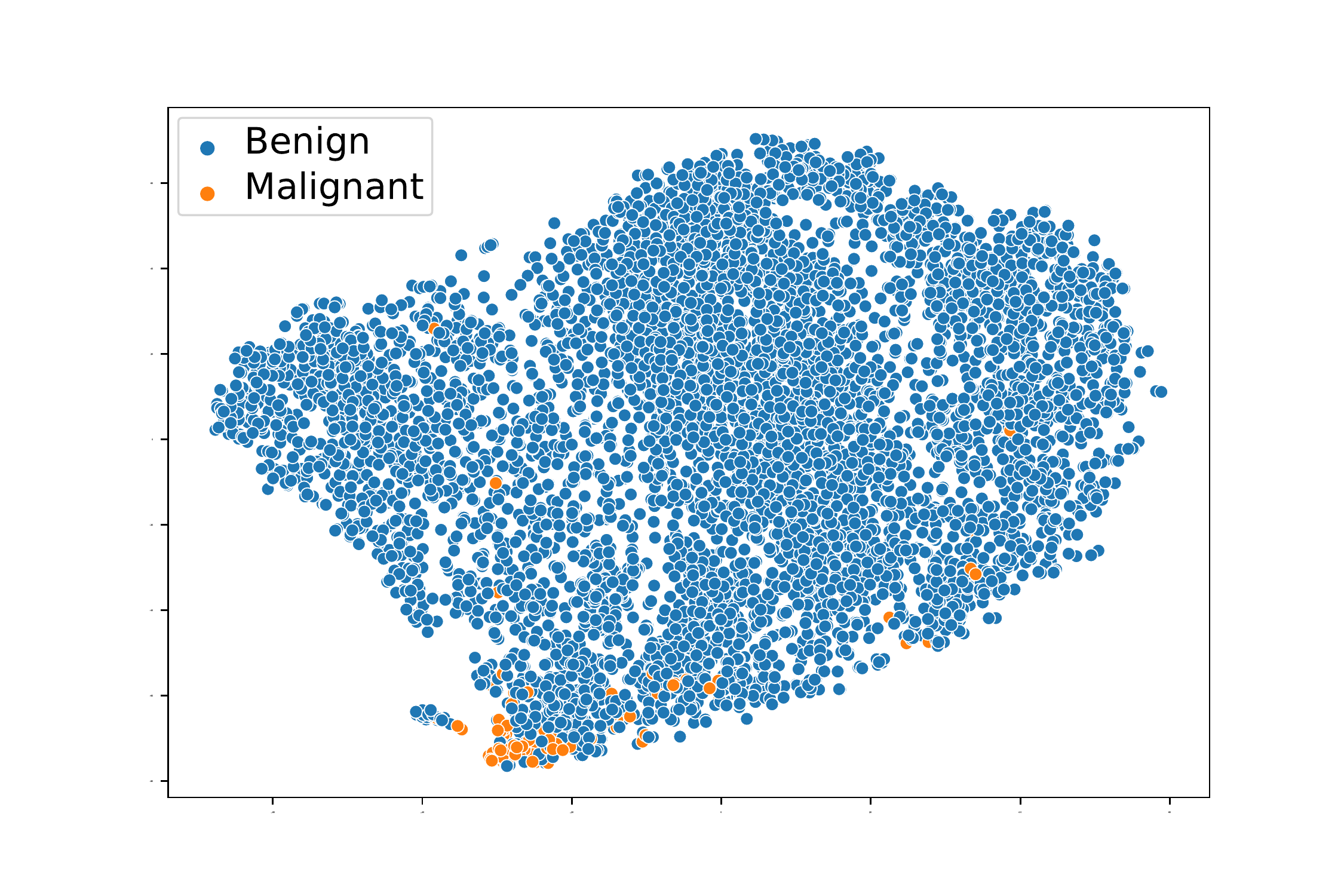}}\hfil \\
\subfloat[]{\label{4}\includegraphics[width=0.33\linewidth]{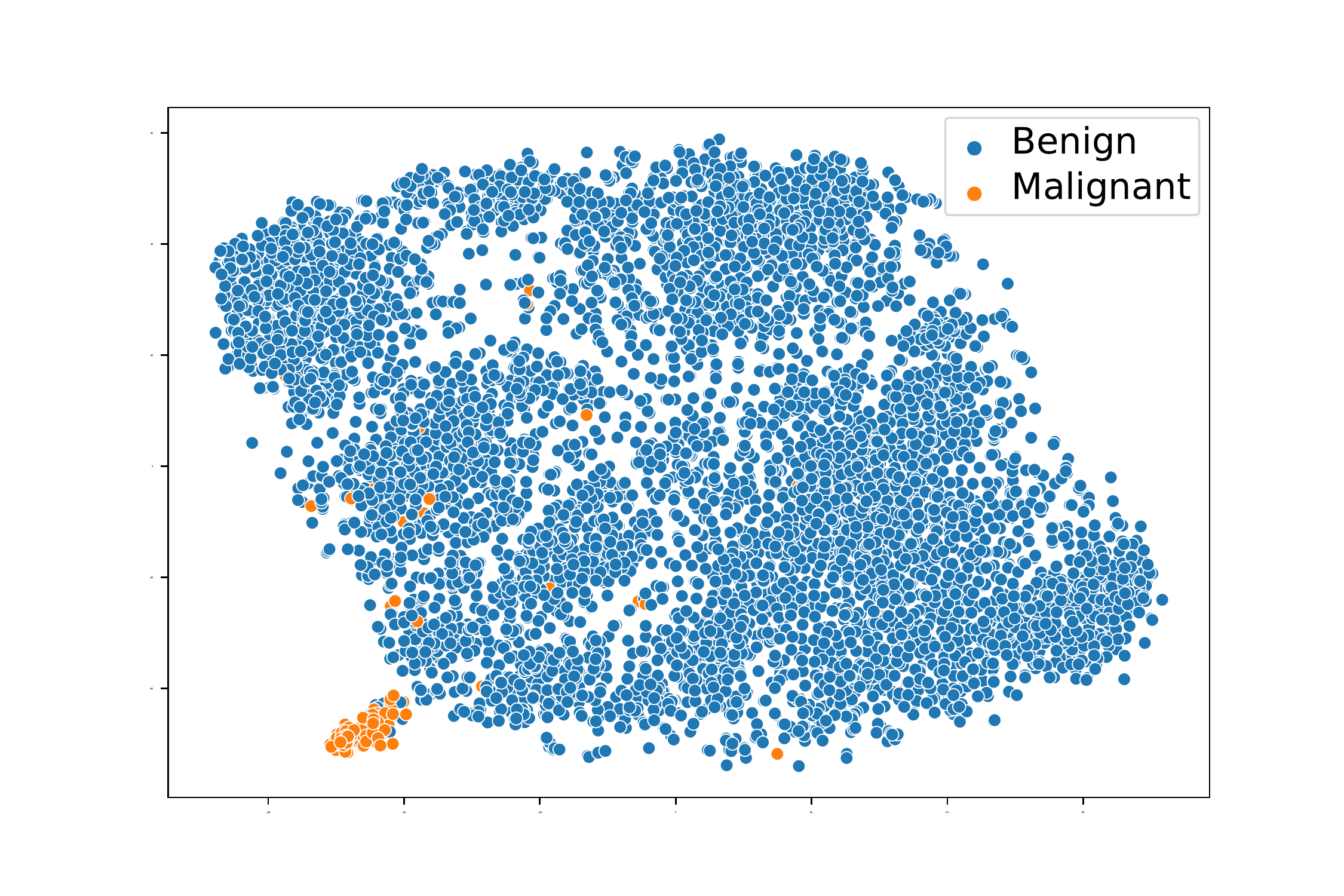}}\hfil
\subfloat[]{\label{5}\includegraphics[width=0.33\linewidth]{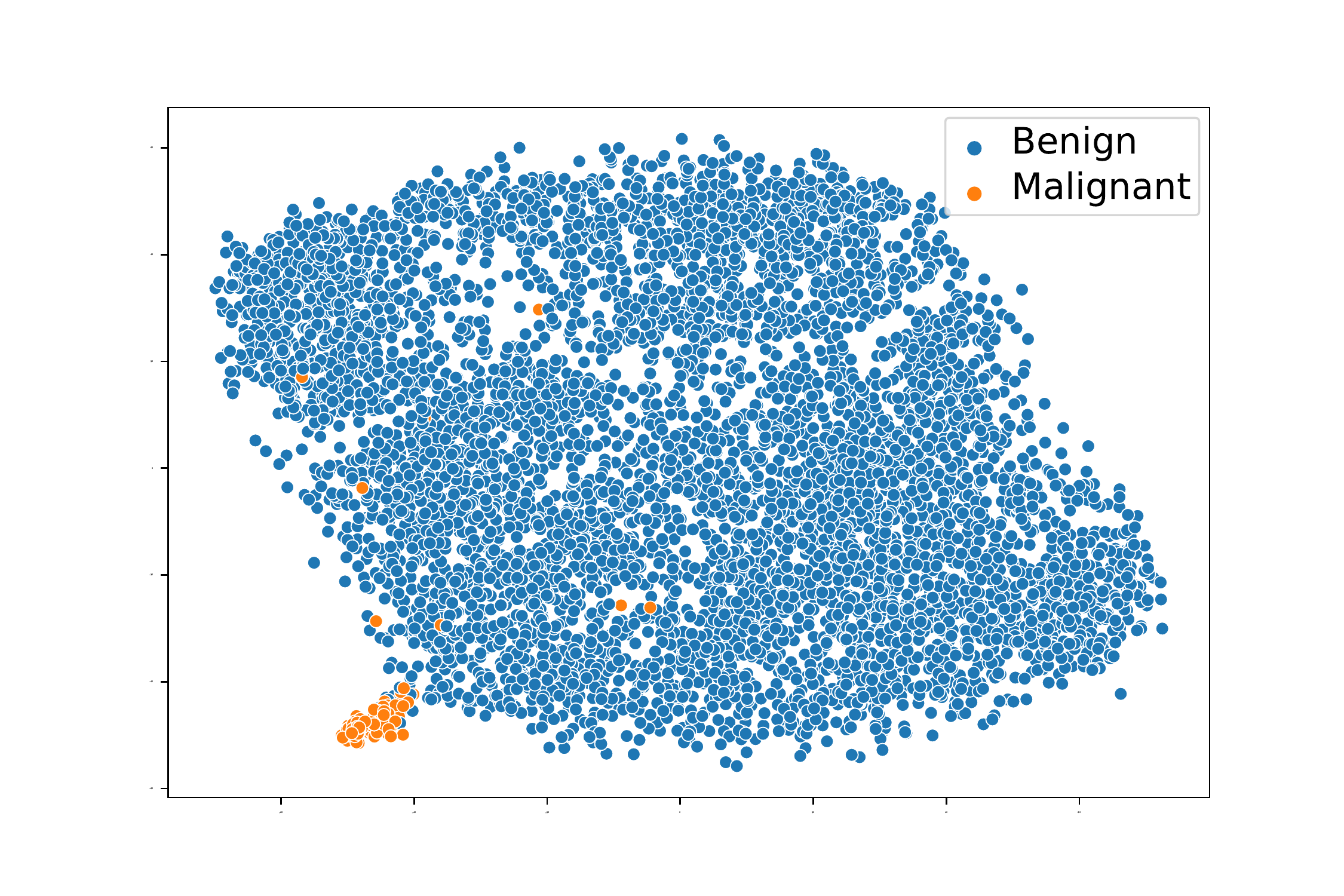}}\hfil
\caption{t-SNE visualization of extracted features from $f_{\theta}$, using (a) Vanilla; (b) Focal-Loss; (c) ROS; (d) SuperCon-CE; (e) SuperCon. Better view in color.}\label{TSNE_plot}
\end{figure*}

\subsection{Evaluation on Extremely imbalanced dataset} \label{exp_2020only}
In the previous section, we conduct a comparison among our SuperCon and other baseline approaches on the dataset ISIC 2019+2020. However, a more imbalanced dataset is more likely to lead failure on conventional training. In this section, we investigate the effectiveness of our two-stage training strategy on more imbalanced dataset. 
To be more specific, we eliminate the additional samples from ISIC 2019 (4,522 melanoma images) and only use the ISIC 2020 on the four ResNets (ResNet18, ResNet50, ResNet101 and ResNet152). In this way, the imbalanced ratio (i.e., the number of majority / the number of minority) is increased significantly from 5.2 to 55.7. Confusion matrix and t-SNE \cite{van2008visualizing} extracted feature visualization of each approach is shown in Figure \ref{confusionmatrix2} and Figure \ref{TSNE_plot}, respectively.
We do not implement RUS \cite{lee2016plankton} under this setting, because the amount of samples (only 467 malignant lesion samples) is not enough for training if we balance the distribution by under-sampling the majority class. One can be considered as a limitation of RUS.

The experimental results are shown in Table \ref{tab2}, we can observe that:
(i) Vanilla consistently has 0.5 on both metrics and Focal-Loss only slightly better than it. 
(ii) ROS has the best AUC among all approaches with ResNet18, but improves limited when using more powerful backbone networks. Its F1-score are the worst and even all lower than 0.5.  
(iii) The performance of SuperCon-CE and SuperCon are much better than other approaches, where they improve the F1-score and AUC by 0.33 and 0.032 in average, respectively. 
(iv) The performance of SuperCon-CE and SuperCon are very close, and the SuperCon consistently better than that without using focal loss on both F1-score and AUC.
(v) Comparing SuperCon-CE and SuperCon in Table \ref{tab1} and \ref{tab2}, the performance with high imbalance ratio are better than that using additional data from ISIC 2019. We consider that the ISIC 2019 and ISIC 2020 are coming from different distributions, although the additional ISIC 2019 and the malignant type in 2020 are cancerous lesions. This could caused by the images are generated by using different devices or different prepocessing protocols. Hence the out of distribution samples misleads the representation training in the first stage, resulting in a lower performance. 

Figure \ref{confusionmatrix2} shows the confusion matrices of the four approaches using ResNet152, where the x-axis and y-axis indicate the predicted and true label. We can see that:
(i) Vanilla simply classifies all sample as benign lesion and none of malignant samples are classified correctly, produces 0 False Positive and 0 True Negative. One is the reason .
(ii) Focal-Loss is able to correctly classify some of the minority class, but the False Positive still very high: $\frac{106}{117} = 90.6\%$.
(iii) ROS shows the best True Negative, but nearly 1/3 majority samples are misclassified.
(iii) Both SuperCon-CE and SuperCon have high True Positive and True Negative, resulting in significant improvement apparently.

We utilize t-SNE \cite{van2008visualizing} to analyze the learnt feature representation of different approaches on ResNet50. We can observe that:
(i) For Vanilla and Focal-Loss, the extracted feature representation from different classes are overlapped and not independent. It means that the model fail to distinguish one from another.
(ii) ROS has a clustered malignant feature representation, but it overlaps with the benign one. This can also explain the high False Positive in Figure. \ref{confusionmatrix2} (c).
(iii) Since both SuperCon-CE and SuperCon have processed the same first stage. They look close to each other and both yield the best separation of different classes.

\subsection{Summary of Experimental Evaluation} \label{discussion}

To summary the experimental evaluation, we investigate the ability of SuperCon to handle the class imbalance on skin lesion dataset. F1-score and AUC score are shown that our SuperCon is outstanding among existing state-of-the-art approaches. Confusion matrices demonstrate that our SuperCon not only yield high True Positives but also high True Negatives, where other approaches hardly reach a high True Negatives or reach with a trade-off (ROS). Extracted feature visualization further illustrates the success using our SuperCon, where an independent and non-overlapping distribution is learnt.
However, SuperCon requires two stages for training, and the performance is sensitive to the number of training epochs in the first stage (Representation training). A proper number of training epochs will obtain a better performance. One can be considered as the limitation of SuperCon.

\section{Conclusion} \label{conclusion}
In this paper, we proposed SuperCon, a two-stage training strategy for class imbalance problem on skin lesion, representation training and classifier fine-tuning. The representation training tries to obtain a feature representation that closely aligned among intra-class and distantly apart from inter-class. While the classifier fine-tuning learns a classifier to address label prediction task on the basis of the obtained representation. Extensive experiments have been conducted to show that SuperCon is able to discriminate different classes under class imbalance settings. The experimental results also show that our SuperCon consistently and significantly beat the existing approaches. For instance, while using a more imbalanced dataset (ISIC 2020), SuperCon outperforms the state-of-the-art Focal-Loss by averaging 0.34 and 0.3, in terms of F1-score and AUC score respectively.
Currently, the number of training epochs in first stage requires manually set based on experience to obtain distinguishable representation. In the future work, we plan to design an end-to-end solution that handles the representation training and classifier fine-tuning simultaneously.

\section*{Acknowledgments} Effort sponsored in part by United States Special Operations Command (USSOCOM), under Partnership Intermediary Agreement No. H92222-15-3-0001-01. The U.S. Government is authorized to reproduce and distribute reprints for Government purposes notwithstanding any copyright notation thereon. \footnote{The views and conclusions contained herein are those of the authors and should not be interpreted as necessarily representing the official policies or endorsements, either expressed or implied, of the United States Special Operations Command.}

\bibliography{mybibfile}

\begin{thebibliography}{10}
\expandafter\ifx\csname url\endcsname\relax
  \def\url#1{\texttt{#1}}\fi
\expandafter\ifx\csname urlprefix\endcsname\relax\def\urlprefix{URL }\fi
\expandafter\ifx\csname href\endcsname\relax
  \def\href#1#2{#2} \def\path#1{#1}\fi

\bibitem{zhang2019attention}
J.~Zhang, Y.~Xie, Y.~Xia, C.~Shen, Attention residual learning for skin lesion
  classification, IEEE transactions on medical imaging 38~(9) (2019)
  2092--2103.

\bibitem{gutman2016skin}
D.~Gutman, N.~C. Codella, E.~Celebi, B.~Helba, M.~Marchetti, N.~Mishra,
  A.~Halpern, Skin lesion analysis toward melanoma detection: A challenge at
  the international symposium on biomedical imaging (isbi) 2016, hosted by the
  international skin imaging collaboration (isic), arXiv preprint
  arXiv:1605.01397 (2016).

\bibitem{codella2018skin}
N.~C. Codella, D.~Gutman, M.~E. Celebi, B.~Helba, M.~A. Marchetti, S.~W. Dusza,
  A.~Kalloo, K.~Liopyris, N.~Mishra, H.~Kittler, et~al., Skin lesion analysis
  toward melanoma detection: A challenge at the 2017 international symposium on
  biomedical imaging (isbi), hosted by the international skin imaging
  collaboration (isic), in: 2018 IEEE 15th International Symposium on
  Biomedical Imaging (ISBI 2018), IEEE, 2018, pp. 168--172.

\bibitem{codella2019skin}
N.~Codella, V.~Rotemberg, P.~Tschandl, M.~E. Celebi, S.~Dusza, D.~Gutman,
  B.~Helba, A.~Kalloo, K.~Liopyris, M.~Marchetti, et~al., Skin lesion analysis
  toward melanoma detection 2018: A challenge hosted by the international skin
  imaging collaboration (isic), arXiv preprint arXiv:1902.03368 (2019).

\bibitem{hensman2015impact}
P.~Hensman, D.~Masko, The impact of imbalanced training data for convolutional
  neural networks, Degree Project in Computer Science, KTH Royal Institute of
  Technology (2015).

\bibitem{lee2016plankton}
H.~Lee, M.~Park, J.~Kim, Plankton classification on imbalanced large scale
  database via convolutional neural networks with transfer learning, in: 2016
  IEEE international conference on image processing (ICIP), IEEE, 2016, pp.
  3713--3717.

\bibitem{pouyanfar2018dynamic}
S.~Pouyanfar, Y.~Tao, A.~Mohan, H.~Tian, A.~S. Kaseb, K.~Gauen, R.~Dailey,
  S.~Aghajanzadeh, Y.-H. Lu, S.-C. Chen, et~al., Dynamic sampling in
  convolutional neural networks for imbalanced data classification, in: 2018
  IEEE conference on multimedia information processing and retrieval (MIPR),
  IEEE, 2018, pp. 112--117.

\bibitem{wang2016training}
S.~Wang, W.~Liu, J.~Wu, L.~Cao, Q.~Meng, P.~J. Kennedy, Training deep neural
  networks on imbalanced data sets, in: 2016 international joint conference on
  neural networks (IJCNN), IEEE, 2016, pp. 4368--4374.

\bibitem{khan2017cost}
S.~H. Khan, M.~Hayat, M.~Bennamoun, F.~A. Sohel, R.~Togneri, Cost-sensitive
  learning of deep feature representations from imbalanced data, IEEE
  transactions on neural networks and learning systems 29~(8) (2017)
  3573--3587.

\bibitem{lin2017focal}
T.-Y. Lin, P.~Goyal, R.~Girshick, K.~He, P.~Doll{\'a}r, Focal loss for dense
  object detection, in: Proceedings of the IEEE international conference on
  computer vision, 2017, pp. 2980--2988.

\bibitem{wang2018predicting}
H.~Wang, Z.~Cui, Y.~Chen, M.~Avidan, A.~B. Abdallah, A.~Kronzer, Predicting
  hospital readmission via cost-sensitive deep learning, IEEE/ACM transactions
  on computational biology and bioinformatics 15~(6) (2018) 1968--1978.

\bibitem{zhang2016training}
C.~Zhang, K.~C. Tan, R.~Ren, Training cost-sensitive deep belief networks on
  imbalance data problems, in: 2016 international joint conference on neural
  networks (IJCNN), IEEE, 2016, pp. 4362--4367.

\bibitem{zhang2018image}
Y.~Zhang, L.~Shuai, Y.~Ren, H.~Chen, Image classification with category centers
  in class imbalance situation, in: 2018 33rd Youth Academic annual conference
  of Chinese Association of Automation (YAC), IEEE, 2018, pp. 359--363.

\bibitem{ren2018multi}
F.~Ren, Y.~Li, M.~Hu, Multi-classifier ensemble based on dynamic weights,
  Multimedia Tools and Applications 77~(16) (2018) 21083--21107.

\bibitem{garcia2018dynamic}
S.~Garc{\'\i}a, Z.-L. Zhang, A.~Altalhi, S.~Alshomrani, F.~Herrera, Dynamic
  ensemble selection for multi-class imbalanced datasets, Information Sciences
  445 (2018) 22--37.

\bibitem{cruz2015meta}
R.~M. Cruz, R.~Sabourin, G.~D. Cavalcanti, T.~I. Ren, Meta-des: A dynamic
  ensemble selection framework using meta-learning, Pattern recognition 48~(5)
  (2015) 1925--1935.

\bibitem{zhuang2020cs}
D.~Zhuang, K.~Chen, J.~M. Chang, Cs-af: A cost-sensitive multi-classifier
  active fusion framework for skin lesion classification, arXiv preprint
  arXiv:2004.12064 (2020).

\bibitem{misra2020self}
I.~Misra, L.~v.~d. Maaten, Self-supervised learning of pretext-invariant
  representations, in: Proceedings of the IEEE/CVF Conference on Computer
  Vision and Pattern Recognition, 2020, pp. 6707--6717.

\bibitem{wu2018unsupervised}
Z.~Wu, Y.~Xiong, S.~X. Yu, D.~Lin, Unsupervised feature learning via
  non-parametric instance discrimination, in: Proceedings of the IEEE
  conference on computer vision and pattern recognition, 2018, pp. 3733--3742.

\bibitem{henaff2020data}
O.~Henaff, Data-efficient image recognition with contrastive predictive coding,
  in: International Conference on Machine Learning, PMLR, 2020, pp. 4182--4192.

\bibitem{sermanet2018time}
P.~Sermanet, C.~Lynch, Y.~Chebotar, J.~Hsu, E.~Jang, S.~Schaal, S.~Levine,
  G.~Brain, Time-contrastive networks: Self-supervised learning from video, in:
  2018 IEEE international conference on robotics and automation (ICRA), IEEE,
  2018, pp. 1134--1141.

\bibitem{khosla2020supervised}
P.~Khosla, P.~Teterwak, C.~Wang, A.~Sarna, Y.~Tian, P.~Isola, A.~Maschinot,
  C.~Liu, D.~Krishnan, Supervised contrastive learning, arXiv preprint
  arXiv:2004.11362 (2020).

\bibitem{tschandl2018ham10000}
P.~Tschandl, C.~Rosendahl, H.~Kittler, The ham10000 dataset, a large collection
  of multi-source dermatoscopic images of common pigmented skin lesions,
  Scientific data 5~(1) (2018) 1--9.

\bibitem{combalia2019bcn20000}
M.~Combalia, N.~C. Codella, V.~Rotemberg, B.~Helba, V.~Vilaplana, O.~Reiter,
  C.~Carrera, A.~Barreiro, A.~C. Halpern, S.~Puig, et~al., Bcn20000:
  Dermoscopic lesions in the wild, arXiv preprint arXiv:1908.02288 (2019).

\bibitem{rotemberg2021patient}
V.~Rotemberg, N.~Kurtansky, B.~Betz-Stablein, L.~Caffery, E.~Chousakos,
  N.~Codella, M.~Combalia, S.~Dusza, P.~Guitera, D.~Gutman, et~al., A
  patient-centric dataset of images and metadata for identifying melanomas
  using clinical context, Scientific data 8~(1) (2021) 1--8.

\bibitem{li2020transformation}
X.~Li, L.~Yu, H.~Chen, C.-W. Fu, L.~Xing, P.-A. Heng, Transformation-consistent
  self-ensembling model for semisupervised medical image segmentation, IEEE
  Transactions on Neural Networks and Learning Systems 32~(2) (2020) 523--534.

\bibitem{yuan2017automatic}
Y.~Yuan, M.~Chao, Y.-C. Lo, Automatic skin lesion segmentation using deep fully
  convolutional networks with jaccard distance, IEEE transactions on medical
  imaging 36~(9) (2017) 1876--1886.

\bibitem{al2018skin}
M.~A. Al-Masni, M.~A. Al-Antari, M.-T. Choi, S.-M. Han, T.-S. Kim, Skin lesion
  segmentation in dermoscopy images via deep full resolution convolutional
  networks, Computer methods and programs in biomedicine 162 (2018) 221--231.

\bibitem{zanddizari2021new}
H.~Zanddizari, N.~Nguyen, B.~Zeinali, J.~M. Chang, A new preprocessing approach
  to improve the performance of cnn-based skin lesion classification, Medical
  \& Biological Engineering \& Computing 59~(5) (2021) 1123--1131.

\bibitem{mishra2019interpreting}
S.~Mishra, H.~Imaizumi, T.~Yamasaki, Interpreting fine-grained dermatological
  classification by deep learning, in: Proceedings of the IEEE/CVF Conference
  on Computer Vision and Pattern Recognition Workshops, 2019, pp. 0--0.

\bibitem{jaworek2019melanoma}
J.~Jaworek-Korjakowska, P.~Kleczek, M.~Gorgon, Melanoma thickness prediction
  based on convolutional neural network with vgg-19 model transfer learning,
  in: Proceedings of the IEEE/CVF Conference on Computer Vision and Pattern
  Recognition Workshops, 2019, pp. 0--0.

\bibitem{barata2019deep}
C.~Barata, J.~S. Marques, M.~Emre~Celebi, Deep attention model for the
  hierarchical diagnosis of skin lesions, in: Proceedings of the IEEE/CVF
  Conference on Computer Vision and Pattern Recognition Workshops, 2019, pp.
  0--0.

\bibitem{perez2019solo}
F.~Perez, S.~Avila, E.~Valle, Solo or ensemble? choosing a cnn architecture for
  melanoma classification, in: Proceedings of the IEEE/CVF Conference on
  Computer Vision and Pattern Recognition Workshops, 2019, pp. 0--0.

\bibitem{noroozi2016unsupervised}
M.~Noroozi, P.~Favaro, Unsupervised learning of visual representations by
  solving jigsaw puzzles, in: European conference on computer vision, Springer,
  2016, pp. 69--84.

\bibitem{he2016deep}
K.~He, X.~Zhang, S.~Ren, J.~Sun, Deep residual learning for image recognition,
  in: Proceedings of the IEEE conference on computer vision and pattern
  recognition, 2016, pp. 770--778.

\bibitem{deng2009imagenet}
J.~Deng, W.~Dong, R.~Socher, L.-J. Li, K.~Li, L.~Fei-Fei, Imagenet: A
  large-scale hierarchical image database, in: 2009 IEEE conference on computer
  vision and pattern recognition, Ieee, 2009, pp. 248--255.

\bibitem{van2008visualizing}
L.~Van~der Maaten, G.~Hinton, Visualizing data using t-sne., Journal of machine
  learning research 9~(11) (2008).

\end{thebibliography}
\end{document}